\documentclass[final,5p,times,twocolumn]{elsarticle}

\usepackage{amssymb,amsfonts,amsmath,amsthm}
\usepackage{graphicx}
\usepackage{microtype}
\usepackage{booktabs}
\usepackage{multirow}
\usepackage{array}
\usepackage{algorithm}
\usepackage{algorithmic}
\usepackage{lineno}
\usepackage{xcolor}
\usepackage{hyperref}
\usepackage{stfloats} 
\usepackage{threeparttable}

\graphicspath{{figures/}} 
\hypersetup{
	colorlinks=true,
	allcolors=blue
}

\DeclareMathOperator{\cov}{cov}

\begin{document}
	\begin{sloppypar}
	\begin{frontmatter}
		
		\title{Trajectory Optimization in Single and Dual-UAV Bearing-Only Target Localization}

		\author[1,2]{Zhijian Xiao}
		\author[1,2]{Huayu Huang}
		\author[1,2,*]{Bin Li}
		\author[1,2,*]{Yang Shang}
		\author[1,2]{Banglei Guan}
		
		\address[1]{College of Aerospace Science and Engineering, National University of Defense Technology, Changsha 410073, China}
		\address[2]{Hunan Key Laboratory of Image Measurement and Visual Navigation, Changsha 410073, China}
		
		\cortext[*]{Corresponding author. \\ \textit{E-mail address:} libin19a@nudt.edu.cn; shangyang1977@nudt.edu.cn.}
		
		\begin{abstract}
			Bearing-only target localization is a fundamental problem in optical measurement and finds extensive applications in unmanned aerial vehicle (UAV) technology. Effective trajectory planning establishes favorable observation geometries, thereby enhancing the target localization accuracy of bearing-only UAV systems. This paper proposes an trajectory optimization method for unmanned aerial vehicles (UAVs) in bearing-only target localization scenarios. By leveraging the Fisher Information Matrix (FIM), the proposed approach dynamically integrates the geometric configuration and vehicle maneuverability into the optimization framework. Specifically, we introduce a spectrally-weighted FIM objective function that provides better gradient dynamics near degenerate configurations, enabling the planner to rapidly escape from poor observation conditions. For dual-UAV scenarios, an intersection angle sine term is introduced to optimize triangulation geometry by improving the sight-line intersection angle, thereby preventing trajectory aggregation. Furthermore, we propose an improved Particle Swarm Optimization (PSO) algorithm with motion model constraints and particle normalization to ensure the physical feasibility of the trajectory and enhance the compatibility with the objective functions. Simulation results demonstrate that the proposed method reduces the median localization error by 99.21\% compared to conventional FIM-based approaches in single-UAV scenarios, and achieves a 69.70\% improvement for dual-UAV configurations, exhibits superior performance in long-duration bearing-only target localization of maneuverability targets at extended ranges.
			
		\end{abstract}
		
		
	\end{frontmatter}
	
	\section{Introduction}
	\label{sec:intro}
	
	Unmanned Aerial Vehicles (UAVs) have widely become indispensable tools in civilian and military applications, including infrastructure inspection, precision agriculture, search and rescue operations, and surveillance missions~\cite{queralta2020collaborative,shule2020uwb}. The ability of UAVs to access areas that are difficult to reach and operate in hazardous environments has driven their rapid adoption in recent years. Central to the effectiveness of these applications is the capability to accurately localize and track targets of interest in real-time, which forms the foundation for autonomous decision-making and mission execution~\cite{demaine2020optimized,zhang2025development}.

    Target localization techniques for UAVs can be categorized based on the type of measurements employed. While range-based methods using GPS, LiDAR, or ultrasonic sensors provide direct distance information, bearing-only localization using monocular cameras has gained significant attention due to its low cost, light weight, and minimal payload requirements~\cite{Pritzl2022cooperative,guo2024research,Luka2025comprehensive}. In most cases, the direction of the targets relative to the UAV is measured through visual sensors, and the position of the targets is estimated through geometric triangulation over time~\cite{he2019bearing}.

    The development of bearing-only localization has been driven by advances in both theoretical understanding and practical algorithms. Early work established the fundamental observability conditions for bearing-only tracking, demonstrating that the observer must execute specific maneuvers to guarantee that the target state is recoverable from angle measurements alone~\cite{nardone1981observability,Aidala1983utilization}. Subsequent research introduced the Fisher Information Matrix (FIM) and Cram\'{e}r-Rao Lower Bound (CRLB) as theoretical tools for quantifying localization accuracy and optimizing sensor placement~\cite{xu2020optimal,xu2019optimal}. These frameworks revealed that localization performance is critically dependent on the relative geometry between observers and targets that means poor viewing angles can lead to unbounded estimation uncertainty even with high-precision sensors~\cite{dogancay2022optimal,aubry2024sensor}.

    Recent advances have significantly expanded the capabilities of bearing-only localization systems. Li et al.~\cite{li2022three} proposed an observability-enhanced helical guidance law that drives UAVs along three-dimensional spiral trajectories, exploiting the additional degree of freedom in altitude to improve observability. Peng et al.~\cite{peng2024trajectory} developed trajectory optimization methods using control barrier functions to actively enhance observability while accounting for sensor bias calibration. Meanwhile, learning-based approaches have emerged as promising alternatives: Fu et al.~\cite{fu2024trajectory} introduced Gaussian process learning for bearing-only target tracking, providing probabilistic bounds on prediction errors. Su et al.~\cite{su2025lowcost} developed a real-time remote sensing framework for uneven terrain using monocular bearing-only micro UAVs, achieving 3.79~m RMSE in field experiments. These developments collectively demonstrate the rich potential of bearing-only methods for practical UAV applications.

    Despite these advances, a fundamental limitation persists: most existing approaches treat the UAV as a passive observer, neglecting its inherent mobility as a controllable degree of freedom for improving localization performance. This observation motivates the paradigm of active perception, where the sensing platform dynamically adjusts its motion to maximize information acquisition~\cite{atanasov2015information}. In contrast to passive data collection with pre-defined trajectories, active perception tightly couples sensing and decision-making, enabling the platform to respond to real-time observations and optimize the quality of gathered information by controlling the motion of the platform~\cite{bajcsy2018revisiting,charrow2015information}.

    The benefits of active perception for bearing-only localization are substantial. By intelligently planning trajectories that maximize observability, UAVs can achieve significantly higher localization accuracy with fewer measurements compared to passive approaches~\cite{hollinger2014sampling,choudhury2020adaptive}. Popovi\'{c} et al.~\cite{popovic2020informative} demonstrated that informative path planning using the Covariance Matrix Adaptation Evolution Strategy (CMA-ES) can reduce map entropy by 45\% compared to coverage-based planners. R\'{u}ckin et al.~\cite{ruckin2022adaptive} showed that deep reinforcement learning for adaptive informative path planning achieves 8--10$\times$ speedup over evolutionary methods while maintaining superior information gathering performance. Recent work by Mudrik et al.~\cite{mudrik2026multiuav} further demonstrated that coordinated teams of fixed-field-of-view UAVs can match or surpass the localization accuracy of single gimballed systems while substantially lowering system complexity and cost. These results highlight the transformative potential of leveraging UAV mobility for active observation.

    However, existing active perception approaches for bearing-only localization face several challenges. First, the conventional FIM-based optimization typically maximizes the determinant~\cite{Mudrik2025optimization}. The D-optimality gradient vanishes near degenerate geometries, causing the optimizer to become trapped in singular configurations. Second, multi-UAV coordination strategies often assume static geometric configurations or decouple trajectory planning from target dynamics~\cite{doostmohammadian2021distributed}. Third, conventional trajectory optimization methods use the direction angle as the optimization vector, resulting in the generation of unreasonable waypoints.~\cite{Qin2023minimum,pan2026spherical}. Recent advances in self-supervised learning for trajectory planning~\cite{Jiang2025selfsupervised} and particle filter-guided neural networks for multi-target tracking~\cite{wang2025particle} offer promising directions. However, their methods have not been fully integrated into the frameworks of active perception.

    To address these limitations, this paper presents an trajectory optimization framework for UAV bearing-only target localization as shown in Fig.~\ref{fig:graphabstract}. Our main contributions are threefold:

    \begin{figure*}[htb]
    \centering
    \includegraphics[width=0.9\textwidth]{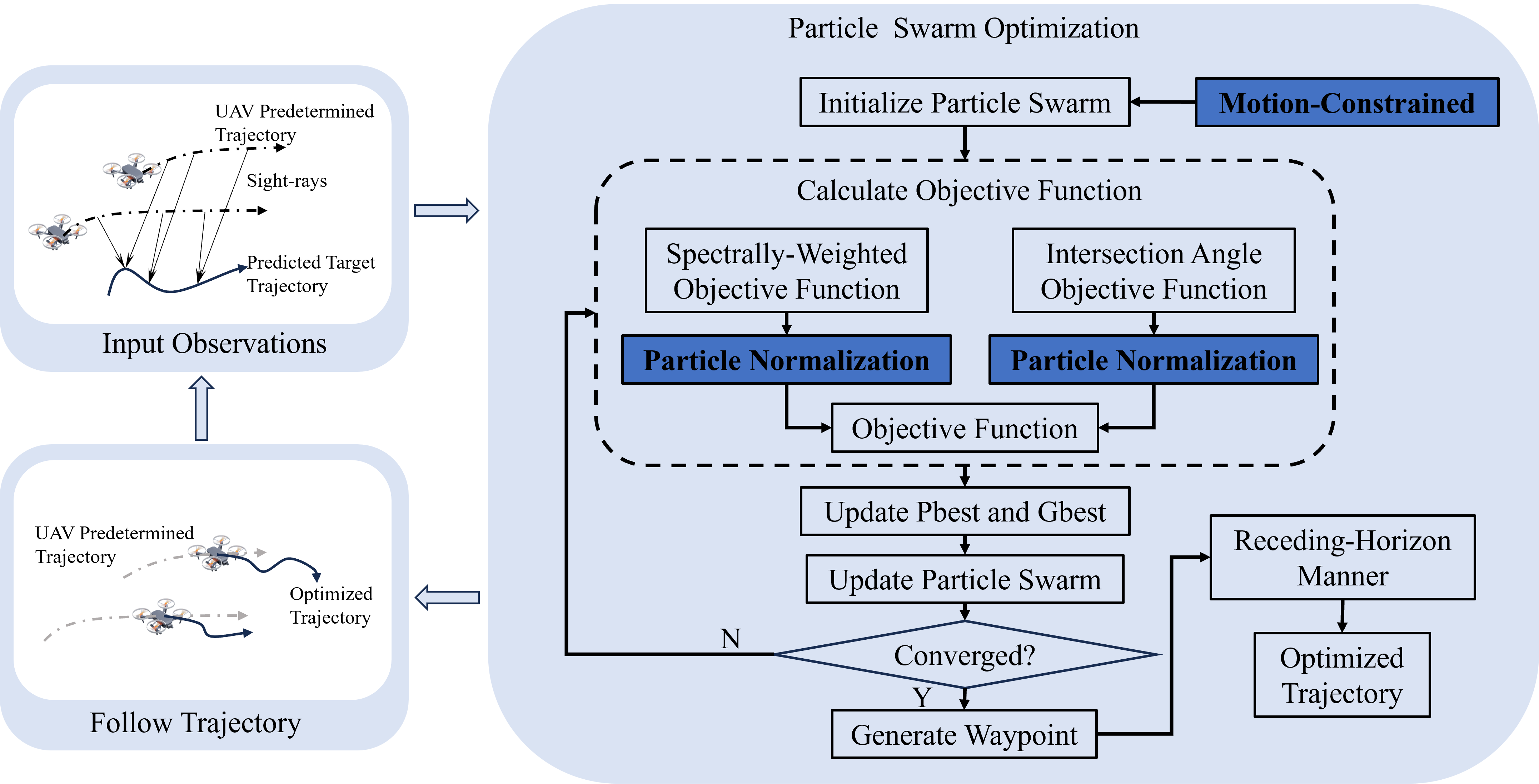}
    \caption{The flow diagram of the proposed method.}
    \label{fig:graphabstract}
    \end{figure*}

\begin{itemize}
    \item We propose a spectrally-weighted FIM objective function that reshapes the optimization landscape near degenerate configurations. By logarithmically aggregating FIM eigenvalues, the objective function generates diverging gradients that enables the optimizer to escape local optima more easily, thereby generating optimal trajectories.
    \item We extend this formulation by introducing an intersection angle sine objective function for dual-UAV scenarios that optimizes triangulation geometry by optimizing the sight-line intersection angle, thereby preventing trajectory aggregation and enhancing cross-bearing information diversity.
    \item We develop a motion-constrained PSO algorithm with particle normalization that embeds platform motion constraints into the particle generation iteration process and uses particle diversity features to automatically adjust the weights of sub-functions. This ensures the physical feasibility of the trajectory without manual tuning of objective weights, enhancing compatibility with the objective functions.
\end{itemize}

    The remainder of this paper is organized as follows. Section~\ref{sec:related} reviews related work on active perception and informative path planning for UAV-based localization. Section~\ref{sec:methods} presents the theoretical foundation of our approach, including the spectrally-weighted FIM formulation, the intersection angle sine objective for dual-UAV coordination, and our optimization algorithm framework. Section~\ref{sec:simulation} presents simulation results validating the effectiveness of the proposed method. Section~\ref{sec:discussion} discusses the obtained results, analyzes the strong and weak points of our method, and outlines future research directions. Section~\ref{sec:conclusion} makes a conclusion for this paper.
	
	\section{Related Work}
	\label{sec:related}
	
	Active perception refers to the paradigm where a sensing platform dynamically adjusts its state or motion to maximize the information content of acquired measurements~\cite{atanasov2015information}. Unlike passive perception where sensor configuration is fixed or follows pre-defined patterns, active perception treats sensing and decision-making as coupled processes that mutually inform each other. This concept has its roots in active vision research, where camera parameters such as gaze direction, focus, and aperture are controlled to facilitate visual tasks~\cite{bajcsy2018revisiting,charrow2015information}. In the context of UAV-based localization, active perception manifests as trajectory optimization that maximizes information-theoretic objectives while respecting platform dynamics and mission constraints.

    The effectiveness of active perception stems from its ability to exploit the fundamental relationship between sensor geometry and estimation uncertainty. For bearing-only measurements, the Fisher Information Matrix provides a theoretical characterization of how observer-target geometry affects localization accuracy~\cite{xu2020optimal,xu2019optimal}. By optimizing trajectories to maximize information content, active perception methods can achieve the same estimation accuracy with significantly fewer measurements compared to passive approaches~\cite{hollinger2014sampling,choudhury2020adaptive}. This efficiency is particularly valuable for UAV operations where battery life and mission duration are constrained.

    Early work in active perception for target localization focused on establishing observability conditions and developing guidance laws that enhance estimation performance. Nardone and Aidala~\cite{nardone1981observability} derived fundamental observability criteria for bearing-only tracking, demonstrating that observer acceleration must satisfy specific conditions to guarantee target state recoverability. Aidala et al.~\cite{Aidala1983utilization} introduced the Fisher Information Matrix for quantifying observability in bearing-only systems, establishing optimal sensor-target geometries that maximize information content. These theoretical foundations laid the groundwork for subsequent trajectory optimization approaches.

    The framework of Informative Path Planning (IPP) formalizes active perception as an optimization problem where trajectories are selected to maximize information-theoretic objectives subject to platform constraints~\cite{popovic2020informative,ruckin2022adaptive}. Within this framework, existing approaches can be broadly classified into two categories: sampling-based and optimization-based methods. Sampling-based algorithms, such as those proposed by Hollinger et al.~\cite{hollinger2014sampling}, balance exploration and exploitation in unknown environments. Choudhury et al.~\cite{choudhury2020adaptive} combined Monte Carlo Tree Search with cost-benefit rollouts to improve sampling efficiency in large action spaces. These methods demonstrated that adaptive replanning based on accumulated information significantly outperforms pre-computed coverage patterns. Optimization-based approaches directly maximize information objectives through continuous trajectory refinement. Popovi\'{c} et al.~\cite{popovic2020informative} employed the Covariance Matrix Adaptation Evolution Strategy (CMA-ES) to optimize measurement positions for UAV-based active classification, achieving substantial reductions in map entropy compared to coverage planners. Hitz et al.~\cite{hitz2017adaptive} extended this approach to persistent monitoring applications, demonstrating continuous refinement of informative trajectories over extended mission durations.

    Recent years have witnessed growing interest in learning-based methods for active perception. R\'{u}ckin et al.~\cite{ruckin2022adaptive} proposed deep reinforcement learning for adaptive informative path planning, training a convolutional neural network to predict informative actions from environmental observations. Their approach achieved 8--10$\times$ speedup over CMA-ES while maintaining superior uncertainty reduction performance. Vashisth et al.~\cite{vashisth2024deep} extended this framework with dynamic graph networks for multi-UAV coordination, enabling scalable adaptive planning through learned communication protocols. Jiang et al.~\cite{Jiang2025selfsupervised} introduced a self-supervised learning approach with differentiable optimization for UAV trajectory planning, achieving 31.33\% improvement in position tracking error and 49.37\% reduction in control effort. These learning-based methods offer promising computational efficiency but require extensive offline training and may struggle with generalization to unseen environments.

    Several approaches specifically address the challenges of bearing-only localization through active perception. He et al.~\cite{he2019bearing} developed trajectory optimization methods that maximize bearing-based observability metrics, revealing that circular motion around the target provides optimal observability for stationary targets. Li et al.~\cite{li2022three} extended this analysis to three-dimensional scenarios, proposing helical guidance laws that exploit altitude variation to enhance observability. Recent work by Zhou et al.~\cite{zhou2026fixedtime} proposed distributed fixed-time estimators for multi-UAV bearing-only target tracking, enabling complete state estimation within preset time bounds. These methods demonstrate the importance of tailoring active perception strategies to the specific characteristics of bearing measurements.

    Multi-UAV active perception introduces additional challenges related to coordination and information fusion. Doostmohammadian et al.~\cite{doostmohammadian2021distributed} developed distributed estimation approaches for tracking mobile targets via UAV formations, incorporating observability analysis into the control design. Dai et al.~\cite{dai2023multi} addressed asynchronous multi-target tracking through collaborative trajectory optimization, demonstrating the benefits of temporal coordination in multi-platform scenarios. Wang et al.~\cite{wang2025energyefficient} proposed energy-efficient cooperative localization by measurement scheduling, demonstrating that optimized fusion strategies can significantly reduce communication costs in large-scale swarms. However, these formation-based and scheduling-centric approaches typically decouple geometric configuration optimization from trajectory planning, limiting their effectiveness when tracking maneuvering targets. The challenge of dynamically optimizing geometric configuration while accounting for target motion and platform constraints remains an open problem.

    In contrast to existing approaches, this work addresses the limitations of current active perception methods for bearing-only localization through three key innovations. First, our spectrally-weighted FIM formulation explicitly promotes robust gradient dynamics near degenerate configurations, overcoming the vanishing-gradient issue of conventional D-optimality criteria. Second, the intersection angle sine objective function integrates geometric configuration planning with trajectory optimization in a unified framework. Third, the motion-constrained PSO with particle normalization ensures the physical feasibility of the trajectory, and enhances compatibility with the objective functions. These contributions collectively advance the state-of-the-art in active perception for UAV-based bearing-only localization.
    
	\section{Methods}
    \label{sec:methods}
    
    In this section, we establish the objective functions for UAV trajectory planning and propose a new algorithmic framework based on PSO algorithm. Section~\ref{sec:fim} introduces the conventional FIM objective function based on the D-optimality criterion and derives its functional expression. Section~\ref{sec:spectral} presents the spectrally-weighted FIM objective function and analyzes its gradient dynamics near degenerate configurations. Section~\ref{sec:intersection} introduces the intersection angle objective for dual-UAV coordination and establishes its equivalence with the GDOP. Section~\ref{sec:pso} presents the optimization algorithm framework for the UAV trajectory planning problem.

    \subsection{Fundamentals of Fisher Information Matrix}
    \label{sec:fim}

    The FIM is a fundamental concept in statistics and information theory, quantifying the sensitivity of probability distributions to parameter variations and serving as a core tool for measuring parameter estimation uncertainty. For a parameterized probability distribution $p(\boldsymbol{z}|\boldsymbol{\theta})$, where $\boldsymbol{z}$ represents the state vector and $\boldsymbol{\theta}$ represents the parameter vector, the FIM is defined based on second-order derivatives of the log-likelihood function. The general expectation-based definition is
    \begin{equation}
    [\mathbf{J}]_{ij} = \mathbb{E}\left[\frac{\partial \ln (p(\boldsymbol{z}|\boldsymbol{\theta}))}{\partial \theta_i} \cdot \frac{\partial \ln (p(\boldsymbol{z}|\boldsymbol{\theta}))}{\partial \theta_j}\right],
    \end{equation}
    where $[\mathbf{J}]_{ij}$ denotes the element in the $i$-th row and $j$-th column of the FIM, and $\frac{\partial \ln p}{\partial \theta_i}$ represents the partial derivative of the log-likelihood function with respect to the $i$-th parameter.

    Diagonal elements of FIM represent the information content for corresponding parameters. Generally, larger diagonal elements indicate stronger sensitivity of the probability distribution to parameter variations, signifying greater parameter importance.

    According to Cram\'{e}r-Rao Lower Bound (CRLB) theory, for an unbiased estimator $\hat{\boldsymbol{\theta}}$, the covariance matrix satisfies
    \begin{equation}
    \cov(\hat{\boldsymbol{\theta}}) \geq \mathbf{J}^{-1}.
    \label{eq:crlb}
    \end{equation}

    Eq.~\eqref{eq:crlb} establishes that the inverse of the FIM constitutes the lower bound for the parameter estimation error covariance matrix. When the observation angle is treated as the observation parameter $\boldsymbol{\theta}$, the target position as the state vector $\boldsymbol{z}$, and the triangulation equation as the probability distribution function $p(\cdot)$ in Eq. 1, the inverse of FIM represents the lower bound of the achievable measurement accuracy. Consequently, researchers often use FIM to assess the impact of observations on estimates. 

    To derive FIM for UAV observations, we simplify the motion of target over a short time interval to uniform acceleration. The state equation for the target is expressed as:
    \begin{equation}
    \boldsymbol{u}(k+1)=\boldsymbol{\Phi}\boldsymbol{u}(k)+\boldsymbol{w},
    \end{equation}
    \begin{equation}
    \boldsymbol{u}(k) = \left[x(k),y(k),z(k),\dot{x}(k), \dot{y}(k), \dot{z}(k), \ddot{x}(k) ,\ddot{y}(k), \ddot{z}(k) \right]^{T},
    \end{equation}
    \begin{equation}
    \boldsymbol{\Phi} = \begin{bmatrix}
    \mathbf{I}_{3} & \text{t}\mathbf{I}_{3} & \dfrac{\text{t}^{2}}{2}\mathbf{I}_{3} \\[10pt]
    \mathbf{0} & \mathbf{I}_{3} & \text{t}\mathbf{I}_{3} \\[10pt]
    \mathbf{0} & \mathbf{0} & \mathbf{I}_{3}
    \end{bmatrix},
    \end{equation}
    where $\boldsymbol{\Phi}$ denotes the state transition matrix, $\boldsymbol{u}(k)$ denotes the state vector of the target at the $k$-th time, $\boldsymbol{w}$ denotes the system process noise, $[x(k),y(k),z(k)]^{T}$ denotes the position of the target at the $k$-th time, and $\text{t}$ denotes the sampling time.

    The observation equation relating the target state to the measured bearings is:
    \begin{equation}
    \begin{aligned}
    \boldsymbol{h} (k) &= \begin{bmatrix} 
    \alpha (k) \\
    \beta (k)
    \end{bmatrix}  \\
    &= \begin{bmatrix} 
    \arctan\left(\frac{y(k)-y_{0}}{x(k)-x_{0}}\right) \\
    \arctan\left(\frac{z(k)-z_{0}}{\sqrt{\left(x(k)-x_{0}\right)^{2} +\left(y(k)-y_{0}\right)^{2}}}\right) 
    \end{bmatrix}+\boldsymbol{\eta},
    \end{aligned}
    \end{equation}
    where $[x_{0},y_{0},z_{0}]^{T}$ denotes the position of the observer at present, $[\alpha(k),\beta(k)]^{T}$ denotes the azimuth and elevation, and $\boldsymbol{\eta}$ denotes the observation noise.

    The FIM matrix for the target observation angle can be calculated using the following formula:
    \begin{equation}
    \label{eq:DiscretizationFIM}
    \mathbf{J}(k) = (\boldsymbol{\Phi}^{-1})^{T}\mathbf{J}(k-1)\boldsymbol{\Phi}^{-1} + \mathbf{H}^{T}(k)\mathbf{R}^{-1}\mathbf{H}(k),
    \end{equation}
    \begin{equation}
    \begin{aligned}
    &\mathbf{H}(k) = \left.\frac{\partial \boldsymbol{h}}{\partial \boldsymbol{u}}\right|_{\hat{\boldsymbol{u}}(k|k-1)}\\
    &= \begin{bmatrix}
    -\dfrac{\sin\alpha}{r_{xy}} &\dfrac{\cos\alpha}{r_{xy}} &0 &\boldsymbol{0}_{1\times6} \\
    -\dfrac{\cos\alpha\sin\beta}{r} &-\dfrac{\sin\alpha\sin\beta}{r} &\dfrac{\cos\beta}{r} &\boldsymbol{0}_{1\times6}
    \end{bmatrix}_{\hat{\boldsymbol{u}}(k|k-1)},
    \end{aligned}
    \end{equation}
    where $\hat{\boldsymbol{u}}(k|k-1)$ represents the predicted state at the $k$-th time, and $r$ and $r_{xy}$ denote the distance of the predicted position of the target relative to the observation platform and its projection onto the horizontal plane.

    Using Eq.~\eqref{eq:DiscretizationFIM}, we can obtain the FIM of the observer's observation of the target at the next time step. As shown in Eq.~\eqref{eq:crlb}, to reduce the observation error, we need to increase the observation FIM. According to the D-optimization criterion, the most common form of FIM quantifier is the determinant form:
    \begin{equation}
    f_{\mathrm{dopt}} = \det(\mathbf{J}).
    \end{equation}

    \subsection{Spectrally-Weighted FIM Objective Function}
    \label{sec:spectral}

    The conventional D-optimality criterion maximizes the determinant of the FIM, which geometrically corresponds to minimizing the volume of the uncertainty ellipsoid. In online trajectory optimization, however, the direct use of the determinant presents practical difficulties when the FIM approaches rank deficiency---a situation that arises whenever multiple sight lines become collinear or nearly collinear. This motivates the adoption of a spectrally-weighted formulation that preserves the fundamental optimality structure while introducing favorable gradient dynamics near such degenerate configurations.

    The spectrally-weighted objective function is formulated as the logarithmic aggregation of FIM eigenvalues:
    \begin{equation}
    \label{eq:spectral_objective}
    f_{\mathrm{spec}}(\boldsymbol{u}) = \sum_{i=1}^{n} \ln\bigl(\lambda_i(\mathbf{J}(\boldsymbol{u}))\bigr),
    \end{equation}
    where $\lambda_i(\mathbf{J}(\boldsymbol{u}))$ denotes the $i$-th eigenvalue of the FIM $\mathbf{J}$ parameterized by the platform motion parameters $\boldsymbol{u}$. Recalling that the determinant of a matrix equals the product of its eigenvalues, this objective function can be rewritten compactly as:
    \begin{equation}
    \label{eq:log_det}
    f_{\mathrm{spec}}(\boldsymbol{u}) = \ln\bigl(\det(\mathbf{J}(\boldsymbol{u}))\bigr) = \ln\bigl(f_{\mathrm{dopt}}(\boldsymbol{u})\bigr),
    \end{equation}

    Because the natural logarithm is a strictly monotone increasing function on $(0,+\infty)$, the ordering of objective function values is preserved under this transformation. Consequently, any parameter vector $\boldsymbol{u}$ that maximizes $f_{\mathrm{dopt}}$ also maximizes $f_{\mathrm{spec}}$, and vice versa. The two objective functions therefore share identical local and global maxima, as well as identical stationary-point structure. The spectrally-weighted formulation does not alter the fundamental optimality properties of the FIM-based trajectory planning problem; rather, its practical advantages stem from the gradient dynamics induced by the logarithmic transformation, which we examine next.

    In bearing-only localization, a \textbf{degenerate (or singular) configuration} occurs whenever multiple sight lines become collinear or nearly collinear. Under such configurations, the FIM loses full rank, its determinant tends to zero, and the corresponding Cram\'{e}r-Rao lower bound diverges, rendering the target state practically unobservable. Both D-optimality and the spectrally-weighted objective function are maximized at the same non-degenerate geometry. The critical difference lies in how their gradients behave when the platform trajectory approaches such degenerate situation.

    To see this, differentiate Eq.~\eqref{eq:log_det} with respect to the motion parameters using the chain rule:
    \begin{equation}
    \label{eq:gradient_chain}
    \nabla f_{\mathrm{spec}}(\boldsymbol{u}) = \frac{1}{\det(\mathbf{J}(\boldsymbol{u}))} \, \nabla f_{\mathrm{dopt}}(\boldsymbol{u}).
    \end{equation}
    
    Taking norms on both sides yields the gradient magnitude ratio as the following:
    \begin{equation}
    \label{eq:gradient_ratio}
    \frac{\|\nabla f_{\mathrm{spec}}(\boldsymbol{u})\|}{\|\nabla f_{\mathrm{dopt}}(\boldsymbol{u})\|} = \frac{1}{\det(\mathbf{J}(\boldsymbol{u}))}.
    \end{equation}

    Equation~\eqref{eq:gradient_ratio} reveals a fundamental behavioral difference. Near a degenerate configuration, $\det(\mathbf{J}) \to 0^{+}$, so the right-hand side diverges to $+\infty$. In other words, the spectrally-weighted objective function generates a gradient whose magnitude scales inversely with the determinant of the FIM. As the observation geometry approaches singularity, this gradient grows without bound, producing a strong \textbf{repulsive force} that pushes the optimization away from the degenerate locus. By contrast, the D-optimality gradient $\nabla f_{\mathrm{dopt}}$ remains bounded (in fact, it tends linearly to zero as the configuration approaches degeneracy), offering no comparable escape mechanism.

    To make this concrete, consider the FIM of a dual-UAV system as a function of the intersection angle $\theta$ between two sight lines. When the UAVs are positioned such that $\theta \to 0$ (collinear sight lines), the FIM determinant behaves asymptotically as $\det(\mathbf{J}) \sim C\theta^{2}$ for some positive constant $C$ determined by the range and measurement noise variance. Substituting this into Eq.~\eqref{eq:gradient_ratio} gives:
    \begin{equation}
    \frac{\|\nabla f_{\mathrm{spec}}\|}{\|\nabla f_{\mathrm{dopt}}\|} \sim \frac{1}{C\theta^{2}} \to +\infty \quad \text{as } \theta \to 0.
    \end{equation}
    
    The spectrally-weighted gradient therefore diverges as $1/\theta^{2}$ near collinearity, whereas the D-optimality gradient vanishes as $\theta$. This disparity is not a matter of numerical conditioning; it is an intrinsic property of the logarithmic transformation.

    The degeneracy-avoidance property translates directly into three operational benefits for online UAV trajectory planning.

    \textbf{(1) Automatic observability safeguarding.} During trajectory optimization, if a candidate motion direction would cause the UAV's sight line to align with a previously recorded bearing, the FIM approaches rank deficiency and $\det(\mathbf{J})$ shrinks. Under D-optimality, the resulting gradient would be too weak to steer the optimizer away from this hazardous direction. The spectrally-weighted objective function, by virtue of Eq.~\eqref{eq:gradient_ratio}, generates an increasingly strong counter-gradient that repels the planned trajectory from collinear geometries. This objective function acts as an implicit observability safeguard without requiring explicit enforcement of minimum bearing-separation angles or hard constraints on angular diversity.

    \textbf{(2) Robust initialization.} In practice, the UAV may begin its mission with a suboptimal initial trajectory---for example, a straight-line approach that keeps the target bearing nearly constant. Such initial segments are close to degenerate, so the D-optimality objective function contributes negligible gradient information, effectively freezing the optimization in its early iterations. Because the spectrally-weighted gradient scales inversely with $\det(\mathbf{J})$, it maintains substantial magnitude precisely in these near-degenerate regions, allowing the planner to correct poor initial geometries rapidly rather than waiting for the UAV to drift into a more favorable configuration by chance.

    \textbf{(3) Stabilized multi-objective function interactions.} When the FIM term is combined with multi-UAV coordination objective functions (addressed in Sections~\ref{sec:intersection} and \ref{sec:pso}), the dynamic range of the D-optimality objective function becomes a practical concern. Near-degenerate flight segments can drive $\det(\mathbf{J})$ down to $10^{-6}$ or lower, whereas favorable geometries may yield values of $10^{4}$ or higher. Spanning ten orders of magnitude within a single sub-objective function destabilizes the relative weighting among competing terms. The logarithmic transformation compresses this range into a moderate interval (roughly $-14$ to $+9$ in natural-log units), ensuring that the FIM term contributes stable gradient information across the entire planning horizon without swamping or being swamped by other sub-objective functions.

    It is worth emphasizing that the advantages described above do not arise from any difference in the global optima of the two formulations. As established by Eq.~\eqref{eq:log_det}, the spectrally-weighted and D-optimality objective functions share the same optimal geometry. The benefits are instead \textit{dynamical} in nature: the logarithmic transformation reshapes the optimization landscape near degenerate boundaries, providing stronger repulsive forces and more stable multi-objective function interactions during the finite-iteration trajectory optimization process.

    \subsection{Intersection Angle Objective Function Extended for Dual-UAV Configuration}
    \label{sec:intersection}

    When multiple observation platforms are deployed, the geometric configuration between the UAVs fundamentally determines the triangulation accuracy. For a dual-UAV bearing-only system, the critical geometric parameter is the \textbf{intersection angle} $\theta$ between the two sight lines at the target location. Intuitively, when the two UAVs view the target from orthogonal directions, the bearing measurements provide complementary information along independent axes, yielding the most reliable position estimate. Conversely, when the sight lines are nearly parallel ($\theta \to 0$), the two measurements become redundant, the triangulation geometry collapses, and the target position uncertainty grows without bound. This angular dependence is therefore central to the design of a multi-UAV objective function.
    
    \begin{figure}[htb]
    \centering
    \includegraphics[width=0.4\textwidth]{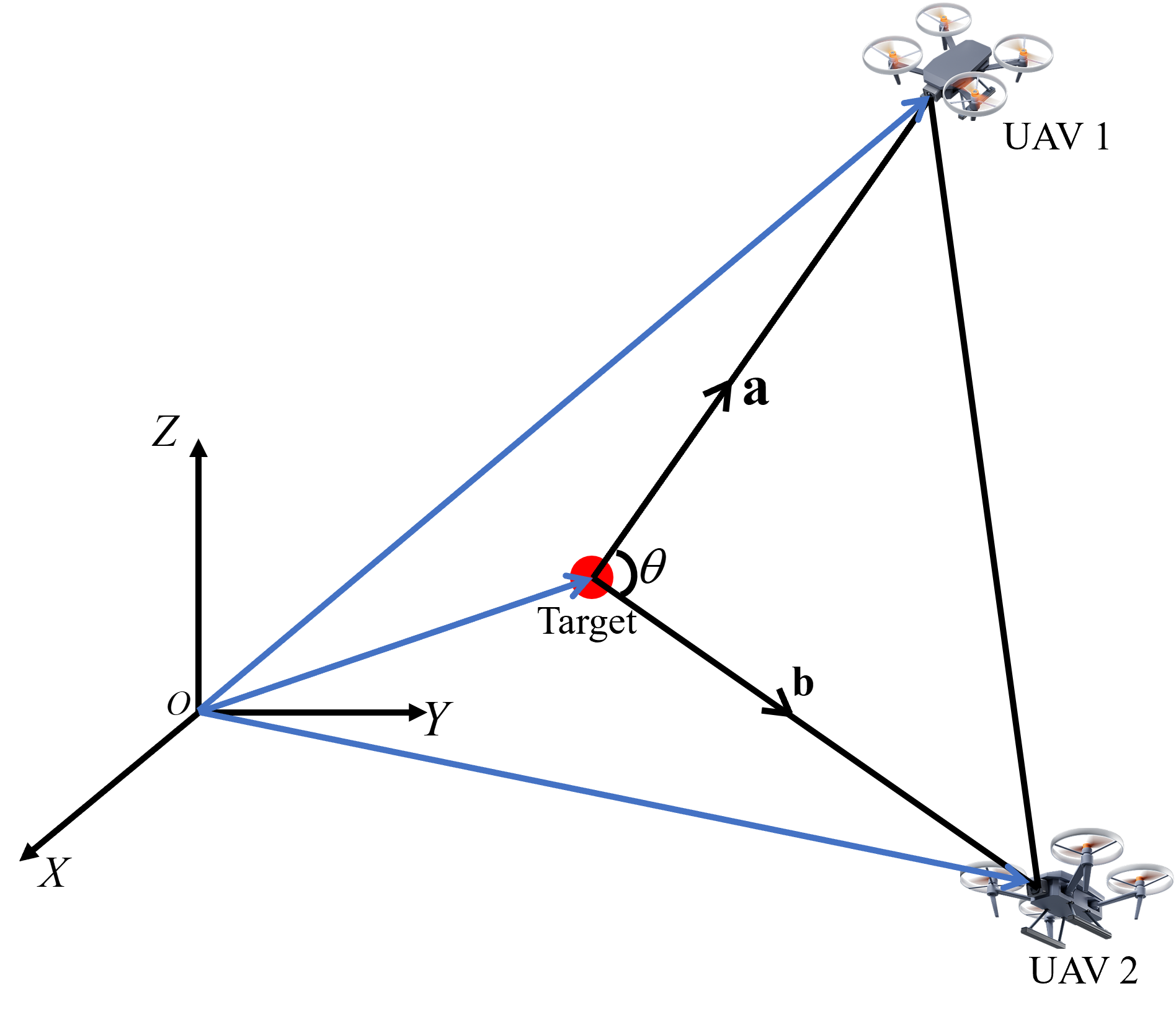}
    \caption{Definition of intersection angle in dual-UAV scenario.}
    \label{fig:intersection}
    \end{figure}

    Consider two observation platforms at positions $\boldsymbol{s}_1, \boldsymbol{s}_2 \in \mathbb{R}^3$ measuring bearing vectors to a target at $\boldsymbol{p} \in \mathbb{R}^3$. Let $\boldsymbol{a} = \boldsymbol{s}_2 - \boldsymbol{p}$ and $\boldsymbol{b} = \boldsymbol{s}_1 - \boldsymbol{p}$ denote the relative position vectors from the target to UAV 2 and UAV 1, respectively. Under equal measurement noise variance $\sigma^2$ for both azimuth and elevation angles, the bearing measurement model yields a Fisher information contribution from each platform that takes the form of a projection matrix orthogonal to the corresponding sight line.

    Specifically, the FIM contribution from a single platform with sight line $\boldsymbol{v}$ is proportional to $\mathbf{P}_{\boldsymbol{v}}^{\perp} = \mathbf{I}_3 - \boldsymbol{v}\boldsymbol{v}^{\top}/\|\boldsymbol{v}\|^2$, which projects onto the plane perpendicular to the line of sight. The total FIM for target position estimation is the sum of contributions from both UAVs as the following:
    \begin{equation}
    \mathbf{J}_{\mathrm{pos}} = \frac{1}{\sigma^2}\left(\mathbf{P}_{\boldsymbol{a}}^{\perp} + \mathbf{P}_{\boldsymbol{b}}^{\perp}\right).
    \end{equation}

    This additive structure reveals that the total information depends on the relative orientation of the two sight lines. When the sight lines are orthogonal, the two projection matrices span complementary subspaces, and the FIM accumulates maximal directional diversity. When the sight lines are parallel, the projection matrices coincide, and the total information degenerates to that of a single observation.

    To quantify this angular dependence, assume for simplicity that both UAVs are at the same distance $d$ from the target, so $\|\boldsymbol{a}\| = \|\boldsymbol{b}\| = d$. Select a coordinate frame in which the two sight lines are symmetrically disposed about the $z$-axis with intersection angle $\theta$:
    \begin{equation}
    \boldsymbol{a} = \begin{bmatrix} 0 \\ d\sin(\theta/2) \\ d\cos(\theta/2) \end{bmatrix}, \qquad
    \boldsymbol{b} = \begin{bmatrix} 0 \\ -d\sin(\theta/2) \\ d\cos(\theta/2) \end{bmatrix}.
    \end{equation}

    Substituting these into the FIM expression and carrying out the matrix algebra, the position FIM becomes diagonal:
    \begin{equation}
    \begin{aligned}
    \mathbf{J}_{\mathrm{pos}} &= \frac{1}{\sigma^2}\begin{bmatrix}
    2 & 0 & 0 \\
    0 & 1 + \cos\theta & 0 \\
    0 & 0 & 2\sin^2(\theta/2)
    \end{bmatrix} \\
    &= \frac{1}{\sigma^2}\begin{bmatrix}
    2 & 0 & 0 \\
    0 & 1 + \cos\theta & 0 \\
    0 & 0 & 1 - \cos\theta
    \end{bmatrix}.
    \end{aligned}
    \end{equation}

    From this diagonal form, the determinant evaluates directly as the product of the diagonal entries:
    \begin{equation}
    \det(\mathbf{J}_{\mathrm{pos}}) = \frac{4(1+\cos\theta)(1-\cos\theta)}{\sigma^6} = \frac{4\sin^2\theta}{\sigma^6 d^4}.
    \end{equation}

    Since $\sin^2\theta \leq 1$ with equality if and only if $\theta = 90^{\circ}$, the FIM determinant is maximized precisely when the two sight lines are orthogonal. The maximum value is $4/(\sigma^6 d^4)$, achieved at the ideal triangulation geometry.

    The preceding derivation establishes that maximizing the FIM determinant is equivalent to maximizing $\sin^2\theta$. It remains to connect this to a more familiar metric in the navigation and surveying literature: the Geometric Dilution of Precision (GDOP).

    The GDOP for position estimation is defined as the square root of the trace of the inverse FIM:
    \begin{equation}
    \mathrm{GDOP} = \sqrt{\mathrm{tr}(\mathbf{J}_{\mathrm{pos}}^{-1})}.
    \end{equation}

    For the diagonal FIM structure above, the inverse is immediate as the following:
    \begin{equation}
    \mathbf{J}_{\mathrm{pos}}^{-1} = \sigma^2
    \begin{bmatrix}
    1/2 & 0 & 0 \\
    0 & 1/(1+\cos\theta) & 0 \\
    0 & 0 & 1/(1-\cos\theta)
    \end{bmatrix},
    \end{equation}
    from which
    \begin{equation}
    \mathrm{GDOP}^2 = \sigma^2\left(\frac{1}{2} + \frac{1}{1+\cos\theta} + \frac{1}{1-\cos\theta}\right) = \sigma^2\left(\frac{1}{2} + \frac{2}{\sin^2\theta}\right).
    \end{equation}

    This expression makes the geometric dependence explicit. As $\theta \to 0$, $\sin\theta \to 0$ and $\mathrm{GDOP} \to +\infty$, reflecting the collapse of triangulation accuracy under collinear sight lines. Conversely, at $\theta = 90^{\circ}$, $\sin\theta = 1$ and the GDOP attains its minimum value $\sigma\sqrt{5/2}$. Because $\mathrm{GDOP}^2$ is a monotone decreasing function of $\sin^2\theta$ on $(0,\pi)$, maximizing $\sin\theta$ is fully equivalent to minimizing the GDOP, and both are equivalent to maximizing the FIM determinant. These equivalences justify the use of the intersection angle sine as the multi-UAV coordination objective function. So we define the objective function as the following:
    \begin{equation}
    \label{eq:intersection_objective}
    f_{\mathrm{sin}} = \sin\theta = \frac{\|\boldsymbol{a} \times \boldsymbol{b}\|}{\|\boldsymbol{a}\| \cdot \|\boldsymbol{b}\|},
    \end{equation}
    where $\boldsymbol{a}$ and $\boldsymbol{b}$ are the sight-line vectors from the target to the two UAVs. The cross-product formulation is coordinate-free and directly computable from the instantaneous relative positions of the platforms and the target, without requiring any angular decomposition into azimuth or elevation components.

    In the trajectory optimization framework, maximizing $f_{\mathrm{sin}}$ drives the two UAVs toward an orthogonal viewing geometry. This prevents the multi-platform system from collapsing into a degenerate configuration where both observers occupy the same bearing direction---a failure mode that is particularly dangerous during long-duration tracking of high-maneuverability targets, where uncoordinated trajectories can converge onto similar flight paths. By incorporating $f_{\mathrm{sin}}$ as an explicit objective function, the planner ensures that the dual-UAV formation maintains favorable triangulation geometry.

    In summary, under symmetric equal-distance assumptions, maximizing the intersection angle sine $f_{\mathrm{sin}} = \sin\theta$ is equivalent to maximizing the FIM determinant and minimizing the position GDOP. This trifold equivalence provides both a theoretical foundation and a practical metric for real-time multi-UAV coordination in bearing-only target localization.

    \subsection{Motion-Constrained PSO with Particle Normalization}
    \label{sec:pso}

    The objective functions established in Sections~\ref{sec:spectral} and \ref{sec:intersection} are highly non-convex with numerous local optima, and their analytical derivatives are intractable due to the geometric coupling between UAV motion and bearing measurements. Evolutionary algorithms are therefore natural candidates for this problem. Among them, Particle Swarm Optimization (PSO) is particularly well suited because its population-based search explores multiple candidate trajectories simultaneously, and its gradient-free nature avoids the need for explicit derivative computation.

    However, applying conventional PSO directly to UAV trajectory planning encounters two practical obstacles. First, since the traditional methods treat the azimuth and pitch of UAVs as the optimization vectors, randomly generated particles may violate platform maneuverability limits, producing physically infeasible trajectories. Second, the weighted-sum aggregation of sub-objective functions requires tedious manual tuning of coefficients when dealing with multiple sub-objective functions, and suffers from scale disparity among terms: the FIM term can range from $10^{-6}$ in near-degenerate geometries to $10^{4}$ in favorable ones, while the intersection sine is bounded in $[0,1]$. Without normalization, the large terms dominate the gradient and the swarm effectively optimizes only these terms, ignoring the others. The consensus is that manual weighting can solve this problem, but inefficiently.

    We address both issues through a two-stage enhancement built upon standard PSO. The complete optimization procedure is presented in Algorithm~\ref{alg:pso} (The same to PSO algorithm, $c_1$ and $c_2$ denotes learning factor, $r_1$ and $r_2$ denotes stochastic factor, $\boldsymbol{p}^{\mathrm{gbest}}$ denots the best particle value discovered  by the entire population, $\boldsymbol{p}^{\mathrm{best}}_i$ denotes the best value that the $i$-th particle has reached since the beginning of the search). The algorithm operates in a receding-horizon manner: at each planning instant, it receives the current UAV state $(\boldsymbol{p}_0, \boldsymbol{v}_0)$,where $\boldsymbol{p}_0$ presents to the position of the UAV at the last time and $\boldsymbol{v}_0$ presents to the velocity of the UAV at the last time, the current target state estimate $\hat{\boldsymbol{u}}(k)$, and the desired planning horizon. It then generates a population of candidate next-waypoints, evaluates them using the spectrally-weighted FIM objective function and the intersection-angle objective function, and returns the waypoint with the highest information utility. The two key enhancements, motion-constrained initialization and projection and particle normalization, are embedded in the particle generation and fitness evaluation steps.

    \begin{algorithm}[htb]
    \caption{Motion-Constrained PSO with Praticle Normalization}
    \label{alg:pso}
    \begin{algorithmic}[1]
    \REQUIRE Current UAV state $\boldsymbol{s}_1=(\boldsymbol{p}_0, \boldsymbol{v}_0)$; target estimate $\hat{\boldsymbol{u}}(k)$; other-UAV state $\boldsymbol{s}_2$; interval $\text{t}$; swarm size $\text{n}$; iterations $\text{k}$; acceleration range of UAVs $\mathbf{\Omega}_a$; velocity range of particles $\mathbf{\Omega}_v$.
    \ENSURE Optimal waypoint $\boldsymbol{p}^*$.
    \STATE // \textit{Step 1: Motion-constrained initialization}
    \FOR{$i = 1$ \TO $\text{n}$}
    \STATE $\boldsymbol{a}_i \sim \mathcal{U}(\mathbf{\Omega}_a)$; \quad $\boldsymbol{p}_i \leftarrow \boldsymbol{p}_0 + \text{t}\boldsymbol{v}_0 + \frac{1}{2}\text{t}^{2}\boldsymbol{a}_i$
    \STATE $\boldsymbol{v}_i \sim \mathcal{U}(\mathbf{\Omega}_v)$; \quad $\boldsymbol{p}_i^{\mathrm{best}} \leftarrow \boldsymbol{p}_i$
    \ENDFOR
    \STATE // \textit{Step 2: First-generation normalization factors}
    \STATE Evaluate $f_{\mathrm{spec}}$, $f_{\mathrm{sin}}$ at all $\boldsymbol{p}_i$; \quad $M^{(m)} \leftarrow \max_i |f_i^{(m)}|$ for $m = \mathrm{spec},\mathrm{sin}$
    \STATE // \textit{Step 3: Iterative evolution}
    \FOR{$j = 1$ \TO $\text{k}$}
    \FOR{$i = 1$ \TO $\text{n}$}
        \STATE $F_i \leftarrow \sum_{m} w_m \frac{f_i^{(m)}}{M^{(m)}}$; \quad update $\boldsymbol{p}_i^{\mathrm{best}}$ and $\boldsymbol{p}^{\mathrm{gbest}}$
        \STATE $\boldsymbol{v}_i \leftarrow w\boldsymbol{v}_i + c_1r_1(\boldsymbol{p}_i^{\mathrm{best}} - \boldsymbol{p}_i) + c_2r_2(\boldsymbol{p}^{\mathrm{gbest}} - \boldsymbol{p}_i)$
        \STATE $\boldsymbol{p}_i' \leftarrow \boldsymbol{p}_i + \boldsymbol{v}_i$; \quad $\boldsymbol{a}' \leftarrow \frac{2}{\text{t}^2}(\boldsymbol{p}_i' - \boldsymbol{p}_0 - \text{t}\boldsymbol{v}_0)$
        \IF{$\boldsymbol{a}' \notin \mathbf{\Omega}_a$}
            \STATE $\boldsymbol{a}' \leftarrow a_{\max}\frac{\boldsymbol{a}'}{\|\boldsymbol{a}'\|}$; \quad $\boldsymbol{p}_i' \leftarrow \boldsymbol{p}_0 + \text{t}\boldsymbol{v}_0 + \frac{1}{2}\text{t}^{2}\boldsymbol{a}'$
        \ENDIF
        \STATE $\boldsymbol{p}_i \leftarrow \boldsymbol{p}_i'$; \quad re-evaluate $f_{\mathrm{spec}}$, $f_{\mathrm{sin}}$ at $\boldsymbol{p}_i$
    \ENDFOR
    \ENDFOR
    \STATE \textbf{return} $\boldsymbol{p}^* \leftarrow     \boldsymbol{p}^{\mathrm{gbest}}$
    \end{algorithmic}
    \end{algorithm}

    \subsubsection{Motion-Constrained Initialization and Projection}

    The first enhancement concerns physical feasibility. In algorithmic applications, the motion angles of UAVs are typically treated as the search space. While this approach has the advantage of reducing the dimensionality of the search space, it fails to account for the dynamic performance. For UAVs with inertial and actuator limits, waypoints optimized using traditional methods may require instantaneous turns or accelerations that exceed the capabilities, causing the inner-loop controller to fail to track the planned trajectory. Alternatively, such methods may fail to fully utilize the maneuverability of UAVs, resulting in overly tightly spaced waypoints. Algorithm~\ref{alg:pso} addresses this by encoding the kinematic model directly into the initialization as the following (Step~1, line~3):
    \begin{equation}
    \boldsymbol{p}_i = \boldsymbol{p}_0 + \text{t}\boldsymbol{v}_0 + \tfrac{1}{2}\text{t}^{2}\boldsymbol{a}_i,
    \end{equation}
    where the maneuver load $\boldsymbol{a}_i$ is sampled from the admissible interval $\mathbf{\Omega_a}$ rather than the position space itself. This guarantees by construction that every particle represents a waypoint reachable within one sampling interval under the actual motion capabilities of the platform.

    \begin{figure}[htb]
    \centering
    \includegraphics[width=0.4\textwidth]{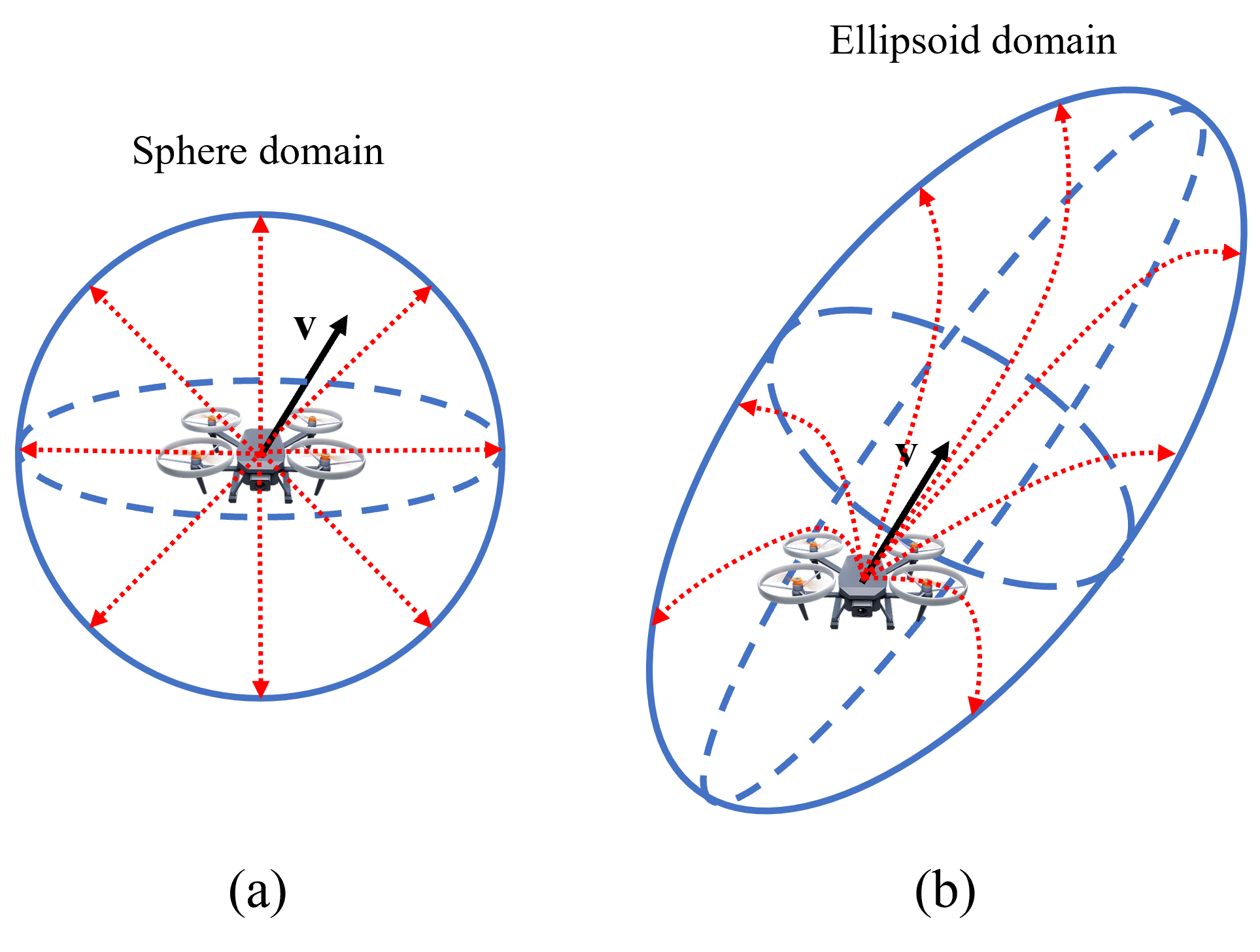}
    \caption{Feasible motion domains: (a) spherical domain assuming omnidirectional agility; (b) ellipsoidal domain considering velocity and acceleration constraints.}
    \label{fig:motion}
    \end{figure}

    Geometrically, the set of all such positions forms an ellipsoidal feasible domain (as illustrated in Fig.~\ref{fig:motion}), whose principal axes are determined by the velocity vector and the acceleration bounds. The spherical feasible region assumed in conventional PSO (which implicitly presumes omnidirectional agility) is replaced by this physically grounded ellipsoid.

    The same motion model is enforced again during swarm evolution (Step~3, lines~9--12). When a particle's velocity update proposes a new position $\boldsymbol{p}_i^{\prime}$, the algorithm computes the acceleration required to reach it from the current state in time $\text{t}$:
    \begin{equation}
    \boldsymbol{a}^{\prime} = \frac{2}{\text{t}^2}(\boldsymbol{p}_i^{\prime} - \boldsymbol{p}_0 - \text{t}\boldsymbol{v}_0).
    \end{equation}
    
    If $\boldsymbol{a}^{\prime}$ exceeds the platform limit $\mathbf{\Omega_a}$, the acceleration vector is clipped to the maximum magnitude $(\boldsymbol{a}_{max})$ and the position is projected back onto the ellipsoidal boundary. This projection step ensures that every iterate (not merely the initialization) remains dynamically feasible. The swarm therefore explores only the physically attainable subset of the search space, preventing the optimizer from wasting evaluations on unreachable waypoints and ensuring that the returned trajectory can be executed by the autopilot without replanning or saturation.

    \subsubsection{Particle Normalization for Adaptive Weighting}

    The second enhancement addresses the multi-objective function aggregation problem. Trajectory planning must balance information acquisition (via $f_{\mathrm{spec}}$) and, in dual-UAV scenarios, triangulation geometry (via $f_{\mathrm{sin}}$). A conventional weighted sum function is defined as the following:
    \begin{equation}
    f_{\mathrm{total}}(\boldsymbol{u}) = \sum_{m} w_m f_m(\boldsymbol{u}), \qquad m = \mathrm{spec},\mathrm{sin}.
    \end{equation}
    where $w_m$ represents the weight coefficient, and satisfies $\sum_{m} w_m = 1$. 
    
    This function suffers from two defects. First, as noted above, the magnitudes of different sub-objective functions may differ by orders of magnitude. Without normalization, the largest term dominates the gradient and the swarm effectively optimizes only that term, ignoring the others. Second, fixing constant weights across the entire mission is suboptimal because the relative importance of information versus geometric coordination evolves as the target maneuvers and the UAV-target geometry changes.

    Algorithm~\ref{alg:pso} resolves both issues through particle normalization (Step~2 and Step~3, line~6). During the first generation, the algorithm records the maximum absolute value  observed for each sub-objective function across the entire swarm:
    \begin{equation}
    M^{(m)} = \max_{i \in [1,N]} |f_m(\boldsymbol{p}_i)|, \qquad m = \mathrm{spec},\mathrm{sin}.
    \end{equation}
    
    These maxima serve as per-generation normalization denominators. The fitness of the $i$-th particle in generation $k$ is then computed as the following:
    \begin{equation}
    F_i = \sum_{m} w_m \frac{f_m(\boldsymbol{p}_i)}{M^{(m)}}.
    \end{equation}
    
    By construction, every normalized sub-objective function value lies in [0, 1] (or a similar range, due to the uncertainty in the distribution of particle states), so no single term can overwhelm the others regardless of its raw scale. The swarm therefore receives balanced gradient contributions from both objective functions, preventing the trajectory planner from sacrificing triangulation geometry for raw information gain, or vice versa. It is worth noting that we have retained the weight $w_m$, which facilitates more intuitive manual allocation rather than requiring multiple rounds of debugging, as is the case with traditional methods. Furthermore, we can even build upon this foundation to add more sub-objective functions while still ensuring that the optimization treats all cases equally.

    Moreover, because the normalization denominators are computed from the first-generation population, they adapt to the current operating region. In a long-range observation phase where all particles exhibit small FIM values, the normalized FIM term retains comparable influence to the intersection term. Conversely, during close-proximity tracking where FIM values are large, the same normalization prevents the FIM term from drowning out the geometric coordination objective function. This adaptive balancing is particularly important in dual-UAV scenarios, where the intersection-angle term must remain active to prevent the two platforms from converging onto similar bearing directions even as the individual FIM terms grow large.

    \subsubsection{Integration with Trajectory Planning}

    In the online execution loop, Algorithm~\ref{alg:pso} is invoked at each planning instant with updated state estimates. The returned waypoint $\boldsymbol{p}^*$ serves as the reference for the UAV's low-level controller for the next sampling interval $\text{t}$. At the subsequent planning cycle, the UAV's measured position and velocity replace $(\boldsymbol{p}_0, \boldsymbol{v}_0)$, the target estimator provides an updated $\hat{\boldsymbol{u}}$, and the optimizer runs again. This receding-horizon structure allows the swarm to react to target maneuvers in real time: if the target changes heading abruptly, the intersection-angle and FIM objective functions shift accordingly, and the swarm naturally migrates toward the new optimal geometry.

    In summary, Algorithm~\ref{alg:pso} combines motion-constrained initialization and projection with particle normalization, addressing the two principal limitations of conventional PSO in UAV trajectory planning. The motion constraints ensure that every evaluated waypoint is dynamically feasible, while the particle normalization guarantees balanced optimization across information and geometric objective functions without manual adjustments.

	\section{Simulation}
	\label{sec:simulation}
	
    To validate the effectiveness of the proposed method, MATLAB was employed as the simulation platform to model a scenario involving multiple observation UAVs tracking a flying target. The trajectory planning for the UAVs was conducted using both the conventional FIM-based trajectory planning method and our proposed method. Sections~\ref{sec:single} and \ref{sec:dual} present detailed analyses of the positioning accuracy and axial error distribution for single-UAV and dual-UAV target localization, respectively.
	
	\subsection{Single-UAV Scenario}
	\label{sec:single}
	
	The initial position of the target is set to $(0, 100, 400)$~m with initial velocity $(-8, 0, 0)$~m/s and acceleration $(1, 1, 0.5)$~m/s$^2$, executing parabolic motion. To demonstrate the superiority of trajectory planning methods, the initial position of UAV is set to $(-1000, 100, 900)$~m with initial velocity $(10, 0, 0)$~m/s and acceleration $(1, 2, 0)$~m/s$^2$, also executing parabolic motion with the same Trajectory Order as the target. The setup of the simulation environment is shown in Tab.~\ref{tab:param_single}.
	
	\begin{table*}[htb]
		\centering
		\caption{Simulation parameters for single UAV observer scenario}
		\label{tab:param_single}
		\begin{tabular}{@{}lcccc@{}}
			\toprule
			Entity & Initial Position (m) & Initial Velocity (m/s) & Acceleration (m/s$^2$) & Trajectory Order \\
			\midrule
			Target & $[0, 100, 400]$ & $[-8, 0, 0]$ & $[1, 1, 0.5]$ & 2 \\
			UAV & $[-1000, 100, 900]$ & $[10, 0, 0]$ & $[1, 2, 0]$ & 2 \\
			\bottomrule
		\end{tabular}
	\end{table*}
	
	To simulate observation errors, camera payload pixel extraction error is set to 2 pixels, camera attitude systematic error to $0.1^\circ$, camera attitude random error to $0.05^\circ$, and camera self-localization random error to 0.1 m.

    This section simulates single-UAV observation of moving targets, utilizing bearing-only measurements as the sole information source for target position estimation. The observation frequency of UAV was set to 10~Hz. During the initial 8 seconds, the UAV follows the predetermined trajectory. Subsequently, trajectory planning is performed every 2 seconds, with the UAV executing planned trajectories while continuously estimating the target position in real-time~\cite{huang20253d}.

    The experimental methods are designated as follows for the ablation study: \textbf{D+PSO} denotes the conventional D-optimality FIM with standard PSO (baseline); \textbf{S+PSO} denotes the spectrally-weighted FIM with standard PSO, testing the effect of the spectral weighting objective function in isolation; \textbf{D+IPSO} denotes the conventional D-optimality FIM with the improved PSO (motion constraints and normalization), testing the effect of the enhanced optimizer in isolation; \textbf{S+IPSO} denotes the spectrally-weighted FIM with the improved PSO, representing the full proposed method that integrates all contributions.
	
	\subsubsection{Experimental Setup and Feasibility Verification}
	\label{sec:feas_single}
	
\begin{figure}[htb]
		\centering
		\includegraphics[width=0.49\textwidth]{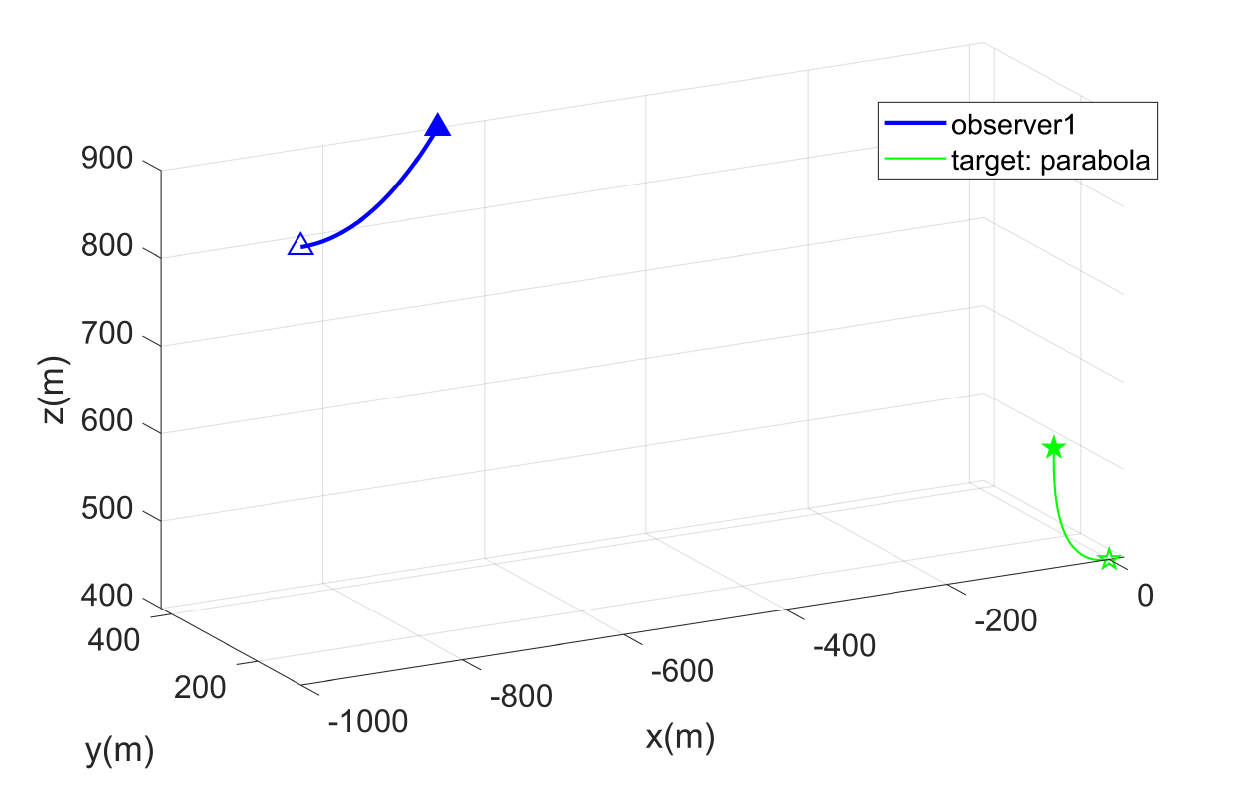}
		\caption{Trajectories of target and single UAV observer.}
		\label{fig:single_tra}
	\end{figure}

\begin{figure}[htb]
		\centering
		\includegraphics[width=0.49\textwidth]{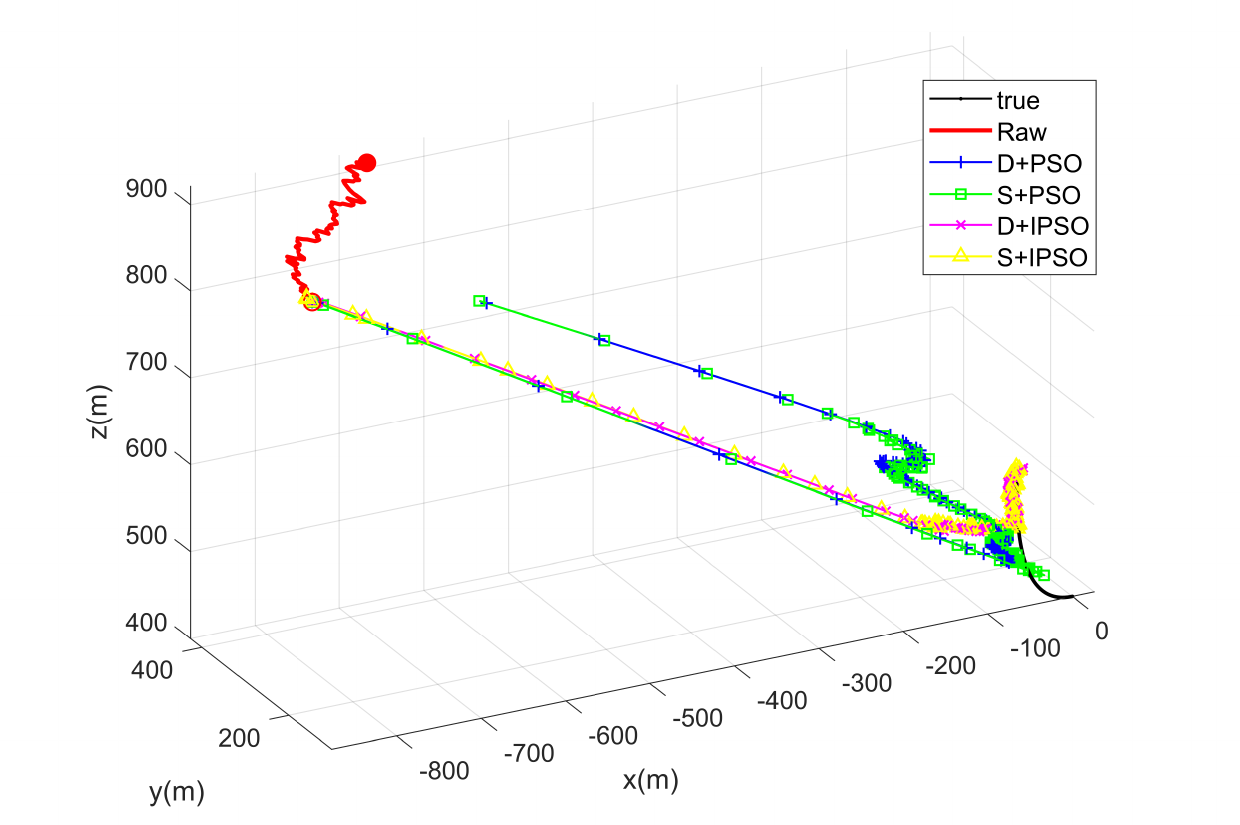}
		\caption{Convergence comparison of target localization.}
		\label{fig:single_result}
	\end{figure}

    Fig.~\ref{fig:single_tra} presents observer and target trajectories, and fig.~\ref{fig:single_result} presents a comparison between the estimated target positions and the actual trajectory for 5 cases:  \textbf{Raw}(No Planning), \textbf{D+PSO}, \textbf{S+PSO}, \textbf{D+IPSO}, and \textbf{S+IPSO} (proposed method).
	
	Evidently, when directly employing the original parabolic trajectory for triangulation-based measurement (No Planning), the estimation erroneously converges to the UAV's own trajectory. This phenomenon may be attributed to the initialization of the UAV's motion parameters as prior values during target position estimation, causing the solution to rapidly converge to the UAV trajectory with minimal iterations based on the observation data. Through trajectory planning, the UAV can promptly escape from this erroneous convergence state and rapidly converge to the true target trajectory, regardless of whether the conventional FIM method or our proposed method is employed. The measurement results during the initial estimation in Fig.~\ref{fig:single_result} approximate the UAV trajectory because the target motion parameter estimates obtained during the unplanned initial phase were used as initialization values. As the estimation process continues, the estimated position gradually converges toward the truth.
	
	\subsubsection{Ablation Study and Accuracy Analysis}
	\label{sec:acc_single}
	
    To validate the accuracy of each component and eliminate the influence of stochastic factors, 100 Monte Carlo simulation trials were conducted. In each trial, the line-of-sight vector errors arising from pixel errors and self-localization errors followed a normal distribution. The measurement results at a specific convergence moment were recorded and statistically analyzed against the ground truth.
	
	\begin{figure}[htb]
		\centering
		\includegraphics[width=0.49\textwidth]{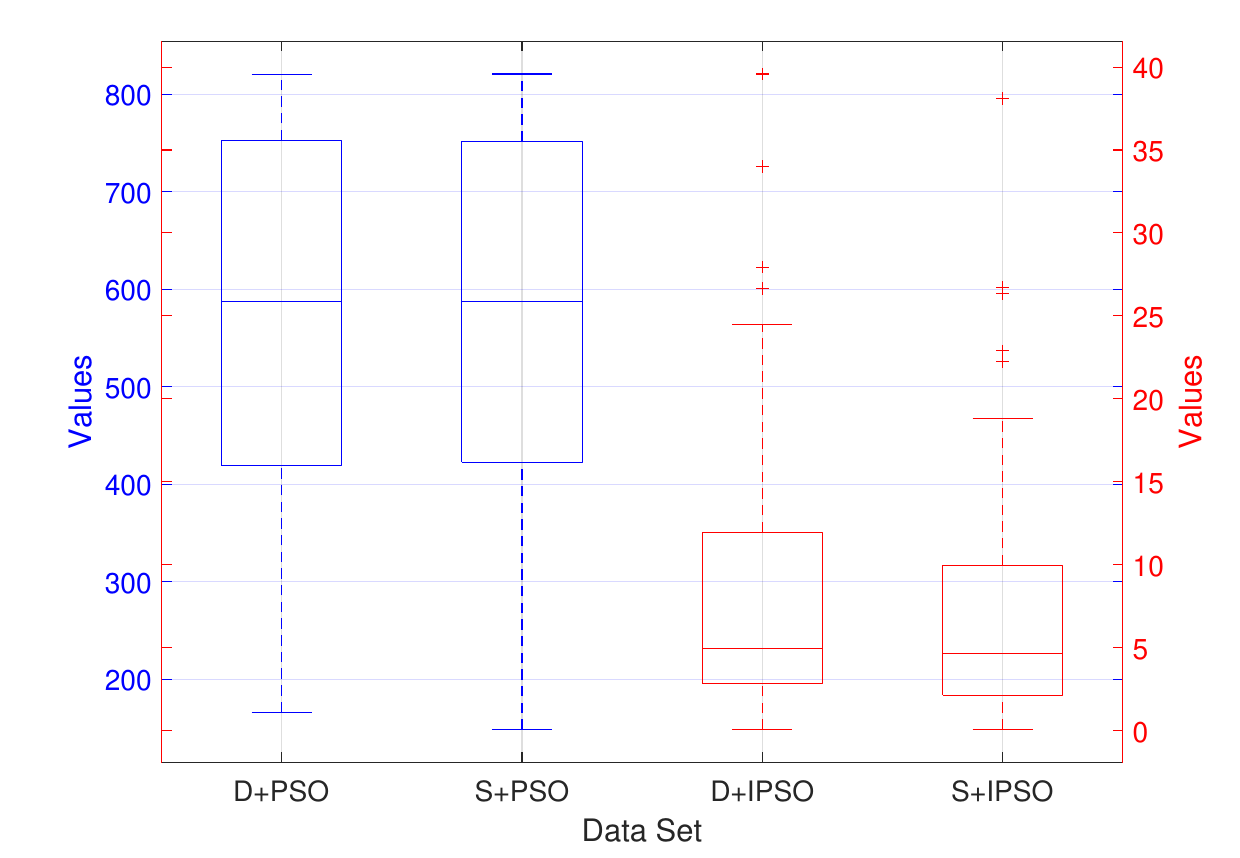}
		\caption{Boxplots of localization errors for the four compared methods (single-UAV scenario).}
		\label{fig:single_box}
	\end{figure}
	
    Fig.~\ref{fig:single_box} represents the statistical distribution of target position measurement errors using boxplots across the four methods. The results demonstrate that the adoption of planned trajectories renders the originally inestimable target position information measurable. The proposed method (\textbf{S+IPSO}) exhibits the most concentrated error distribution with the lowest median and the tightest interquartile range.
	
	\begin{table}[htb]
		\centering
        \small
		\caption{Statistical comparison of localization errors for single-UAV scenario (meters)}
		\label{tab:err_single}
		\begin{tabular}{@{}lcccccc@{}}
			\toprule
			Method & Median & Mean & Std Dev & Outliers & Max & Min \\
			\midrule
			Raw & -- & 815.65 & 32.68 & -- & -- & -- \\
			D+PSO & 587.97 & 548.53 & 218.96 & 0 & 820.39 & 166.36 \\
			S+PSO & 587.82 & 548.49 & 218.44 & 0 & 820.91 & 148.33 \\
            D+IPSO & 4.94 & 8.38 & 8.03 & 5 & 39.58 & 0.06 \\
            S+IPSO & 4.63 & 6.77 & 6.78 & 5 & 38.10 & 0.08 \\
			\bottomrule
		\end{tabular}
	\end{table}
	
    The ablation study results in Table~\ref{tab:err_single} reveal distinct roles of the proposed components. Comparing \textbf{S+PSO} against \textbf{D+PSO}, the replacement of D-optimality with the spectrally-weighted objective function while retaining the standard PSO yields negligible improvement (0.03\% reduction in median error, 0.01\% in mean error). This indicates that the spectrally-weighted formulation alone, without the motion-constrained optimizer, cannot overcome the limitations of conventional PSO in navigating the highly non-convex planning landscape.

    Comparing \textbf{D+IPSO} against \textbf{D+PSO}, the introduction of the improved PSO (motion-model-constrained initialization and adaptive normalization) while retaining the conventional D-optimality criterion produces a dramatic accuracy improvement: the median error is reduced by \textbf{99.16\%} and the mean error by \textbf{98.47\%}. This confirms that the motion-constrained PSO is the dominant contributor to performance gain, as its physically grounded initialization prevents infeasible waypoints and the adaptive normalization stabilizes the optimization process.

    Finally, comparing \textbf{S+IPSO} (proposed method) against \textbf{D+PSO}, the integration of both the spectrally-weighted objective function and the improved PSO achieves the best overall performance, with a \textbf{99.21\%} reduction in median error and a \textbf{98.77\%} reduction in mean error. Relative to D + IPSO, S + IPSO achieves an additional \textbf{6.28\%} median reduction and \textbf{19.21\%} mean reduction, demonstrating that the spectrally-weighted objective function provides further refinement when coupled with the improved optimizer. The spectral weighting reshapes the local optimization landscape near degenerate geometries, allowing the swarm to more efficiently exploit favorable configurations once the motion constraints have brought particles into physically attainable regions.
	
	\subsubsection{Axial Error Distribution Analysis}
	\label{sec:iso_single}
	
\begin{figure*}[htb]
		\centering
		\includegraphics[width=0.9\textwidth]{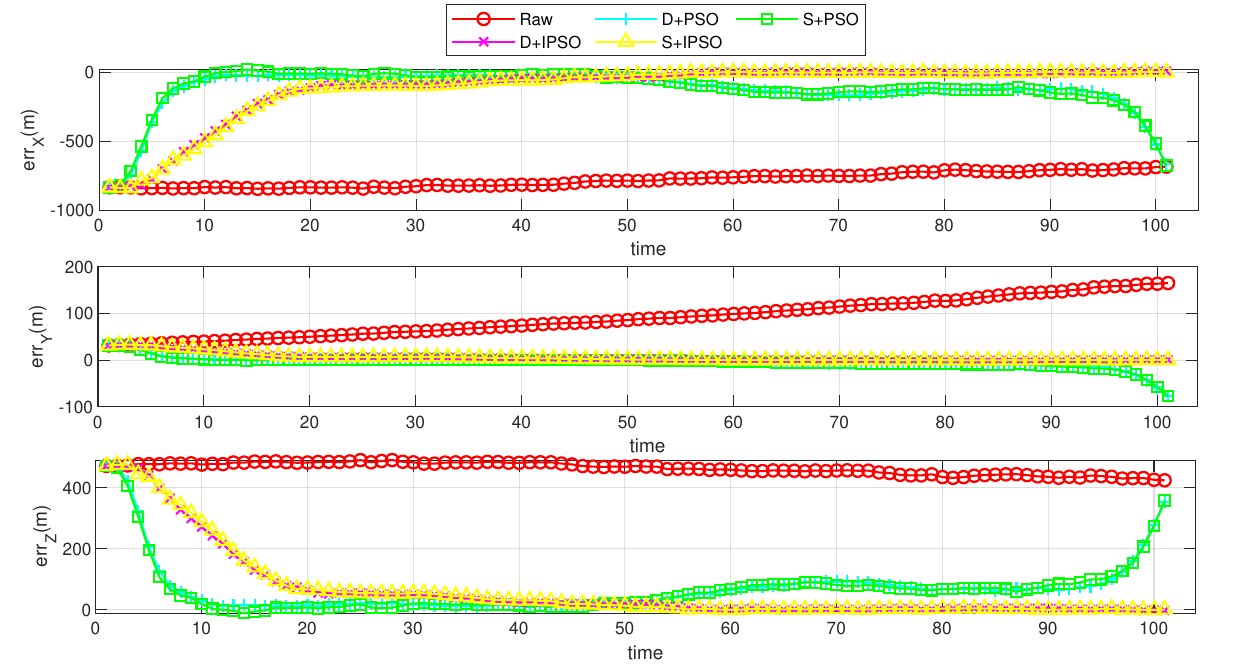}
		\caption{Localization error decomposition along x, y, and z axes for single UAV observer scenario.}
		\label{fig:single_iso}
	\end{figure*}

    To examine how localization errors are distributed across the three spatial dimensions, the target position estimation errors along the $x$, $y$, and $z$ axes were extracted from the simulation results. Fig.~\ref{fig:single_iso} illustrates the localization errors along the $x$, $y$, and $z$ axes during UAV target tracking. The results indicate that, under comparable total localization errors, the proposed method (\textbf{S+IPSO}) yields more balanced error components across all three axes compared to the conventional FIM method (\textbf{D+PSO}), which exhibits strongly axis-dependent behavior.
	
\begin{figure}[tb]
		\centering
		\includegraphics[width=0.49\textwidth]{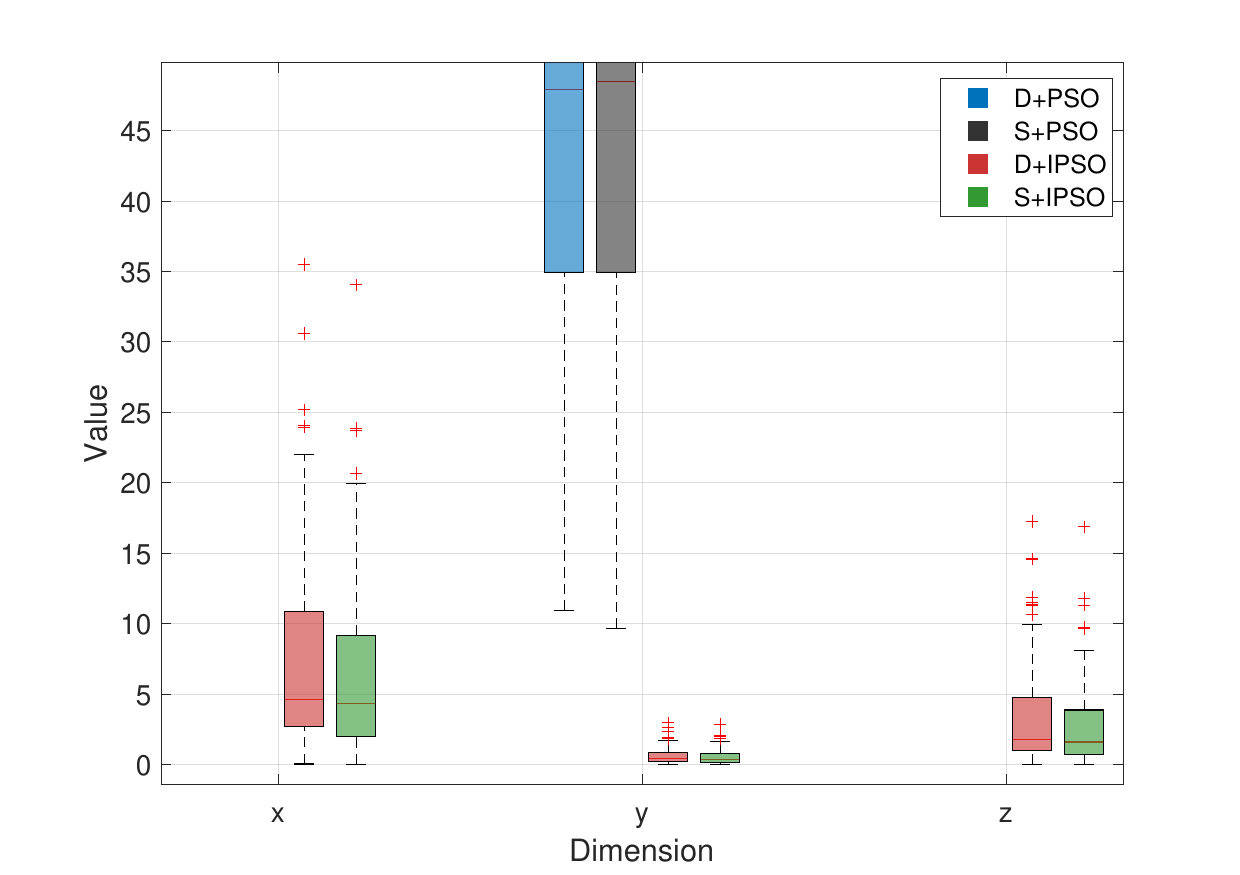}
		\caption{Box plots of localization error components for single UAV observer scenario.}
		\label{fig:single_iso_box}
	\end{figure}

    To mitigate stochastic effects, the error components along the $x$, $y$, and $z$ axes from the 100 simulation trials were statistically analyzed and presented as boxplots. Fig.~\ref{fig:single_iso_box} and Table~\ref{tab:iso_single} demonstrate the differences in the distribution of axial errors.
	
	\begin{table}[htb]
		\centering
		\caption{Axial error distribution comparison for single-UAV scenario (meters)}
		\label{tab:iso_single}
		\resizebox{\columnwidth}{!}{
			\begin{tabular}{@{}lccccccc@{}}
				\toprule
				Axis & Method & Median & Mean & Std Dev & Outliers & Max & Min \\
				\midrule
				\multirow{4}{*}{$x$} 
                & D+PSO & 517.18 & 482.23 & 192.21 & 0 & 721.04 & 147.51 \\
				& S+PSO & 517.04 & 483.19 & 191.76 & 0 & 721.27 & 131.96 \\
                & D+IPSO & 4.62 & 7.63 & 7.19 & 5 & 35.49 & 0.04 \\
                & S+IPSO & 4.34 & 6.16 & 6.08 & 4 & 34.05 & 0.02 \\
				\midrule
                \multirow{4}{*}{$y$} 
                & D+PSO & 47.91 & 48.95 & 23.74 & 0 & 84.29 & 10.91 \\
				& S+PSO & 48.54 & 48.97 & 23.65 & 0 & 84.58 & 9.64 \\
                & D+IPSO & 0.43 & 0.66 & 0.62 & 8 & 2.97 & 0.004 \\
                & S+IPSO & 0.39 & 0.54 & 0.53 & 4 & 2.86 & 0.004 \\
				\midrule
                \multirow{4}{*}{$z$} 
                & D+PSO & 276.24 & 256.70 & 102.39 & 0 & 386.06 & 76.16 \\
				& S+PSO & 276.24 & 256.70 & 102.15 & 0 & 386.63 & 67.04 \\
                & D+IPSO & 1.77 & 3.39 & 3.54 & 6 & 17.26 & 0.02 \\
                & S+IPSO & 1.60 & 2.73 & 2.96 & 5 & 16.87 & 0.02 \\
				\bottomrule
			\end{tabular}
		}
	\end{table}
	
    The conventional FIM method (\textbf{D+PSO}) achieves substantially different accuracy levels along different axes: the median error along the $y$-axis (47.91~m) is an order of magnitude smaller than that along the $x$-axis (517.18~m), reflecting the anisotropic nature of D-optimality when the observer trajectory is predominantly aligned with one direction. In contrast, the proposed method (\textbf{S+IPSO}) reduces errors consistently across all axes. Notably, the $y$-axis accuracy remains the highest (median 0.39~m), while the $x$- and $z$-axis errors are brought to comparable levels (median 4.34~m and 1.60~m, respectively), indicating that our method slightly mitigates the directional bias inherent in conventional determinant maximization.
	
	\subsection{Dual-UAV Scenario}
	\label{sec:dual}
	
	The target initial position is set the same as Section~\ref{sec:single}. Two UAVs' initial positions are set to $(-1000, 100, 900)$~m and $(-1000, 0, 900)$~m, both with initial velocity $(5, 2, 0)$~m/s, executing constant velocity motion (Trajectory Order is 1). The setup of the simulation environment is shown in Tab.~\ref{tab:param_dual}. Observation error parameters are identical to Section~\ref{sec:single}.
    \begin{table*}[htb]
		\centering
		\caption{Simulation parameters for dual-UAV scenario}
		\label{tab:param_dual}
		\begin{tabular}{@{}lcccc@{}}
			\toprule
			Entity & Initial Position (m) & Initial Velocity (m/s) & Acceleration (m/s$^2$) & Trajectory Order \\
			\midrule
			Target & $[0, 100, 400]$ & $[-8, 0, 0]$ & $[1, 1, 0.5]$ & 2 \\
			UAV 1 & $[-1000, 100, 900]$ & $[5, 2, 0]$ & $[0, 0, 0]$ & 1 \\
			UAV 2 & $[-1000, 0, 900]$ & $[5, 2, 0]$ & $[0, 0, 0]$ & 1 \\
			\bottomrule
		\end{tabular}
	\end{table*}

    To simulate observation errors, camera payload pixel extraction error is set to 2 pixels, camera attitude systematic error to $0.1^\circ$, camera attitude random error to $0.05^\circ$, and camera self-localization random error to 0.1 m.
	\begin{figure}[tb]
		\centering
		\includegraphics[width=0.49\textwidth]{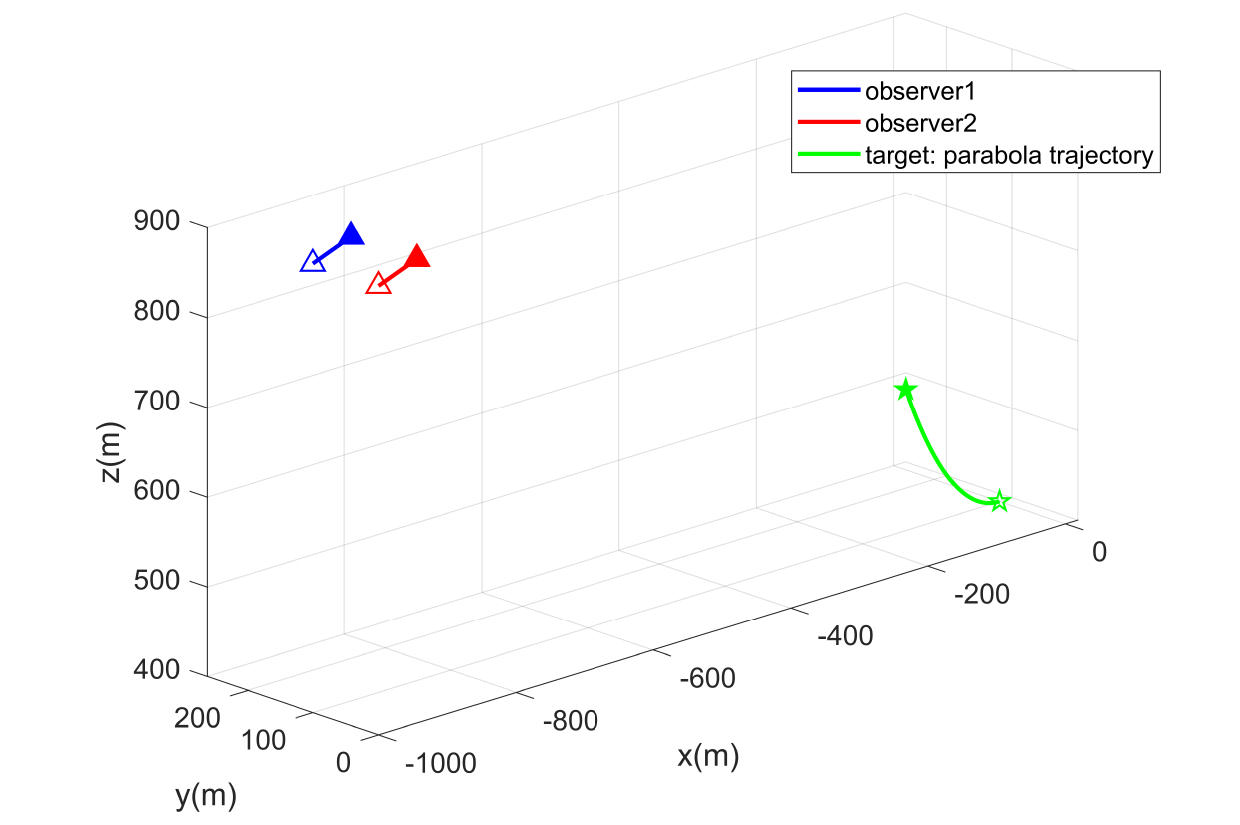}
		\caption{Trajectories of target and dual UAV observers.}
		\label{fig:dual_tra}
	\end{figure}

    This section simulates dual-UAV observation of moving targets, utilizing bearing-only measurements from two platforms for target position estimation. The observation frequency was set to 10~Hz. During the initial 8 seconds, the UAVs followed predetermined trajectories to acquire sufficient target information. Subsequently, trajectory planning was performed every 2 seconds, and the UAVs were controlled to follow the optimized trajectories while continuously estimating the position of the target via triangulation.

    The experimental methods are designated as follows for the ablation study: \textbf{D+PSO} denotes the conventional D-optimality FIM with standard PSO (baseline); \textbf{S+PSO} denotes the spectrally-weighted FIM with standard PSO; \textbf{D+IPSO} denotes the conventional D-optimality FIM with the improved PSO; \textbf{DA+PSO} denotes the D-optimality FIM with the intersection angle sine and standard PSO, testing the geometric coordination objective function in isolation; \textbf{SA+IPSO} denotes the spectrally-weighted FIM with the intersection angle sine and the improved PSO, representing the full proposed method that integrates all contributions.
	\subsubsection{Experimental Setup and Feasibility Verification}
	\label{sec:feas_dual}
    
\begin{figure}[htb]
		\centering
		\includegraphics[width=0.49\textwidth]{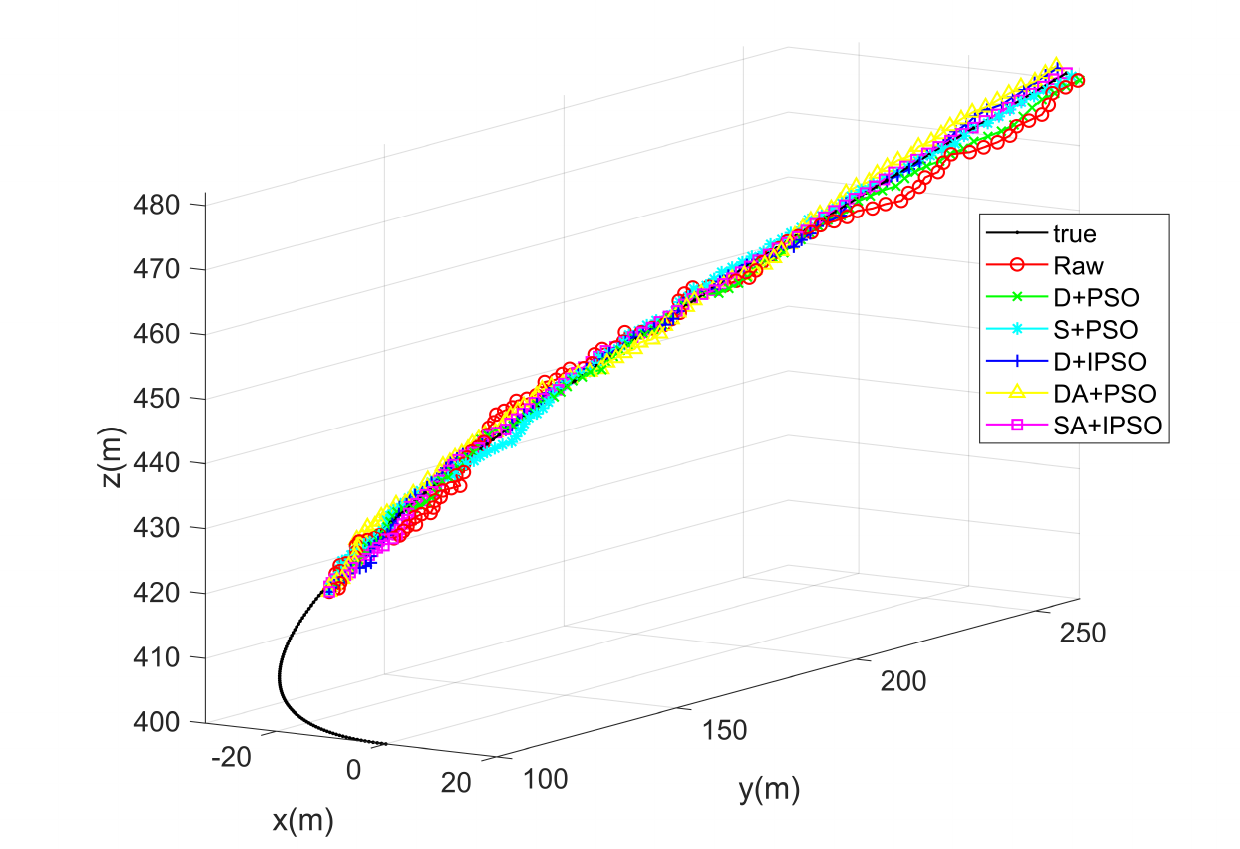}
		\caption{Convergence comparison for dual-UAV scenario.}
		\label{fig:dual_result}
	\end{figure}

    Fig.~\ref{fig:dual_result} presents a comparison between the estimated target positions and the actual trajectory for 6 cases: \textbf{Raw}(No Planning), \textbf{D+PSO}, \textbf{S+PSO}, \textbf{D+IPSO},  \textbf{DA+PSO}, and \textbf{SA+IPSO}.
	
	Unlike the monocular measurement scenario, the dual-UAV configuration enables adequate target information acquisition even when the UAV trajectory order is lower than that of the target motion, yielding relatively reliable target position estimates. Nevertheless, trajectory planning enables the UAVs to achieve superior position solutions using the same triangulation algorithm.
	
	\subsubsection{Ablation Study and Accuracy Analysis}
	\label{sec:acc_dual}
	
    To validate accuracy and eliminate stochastic effects, 100 Monte Carlo simulation trials were conducted. Line-of-sight vector errors followed a normal distribution in each trial. Measurement results at a specific convergence moment were recorded and statistically analyzed.
	
	\begin{figure}[tb]
		\centering
		\includegraphics[width=0.49\textwidth]{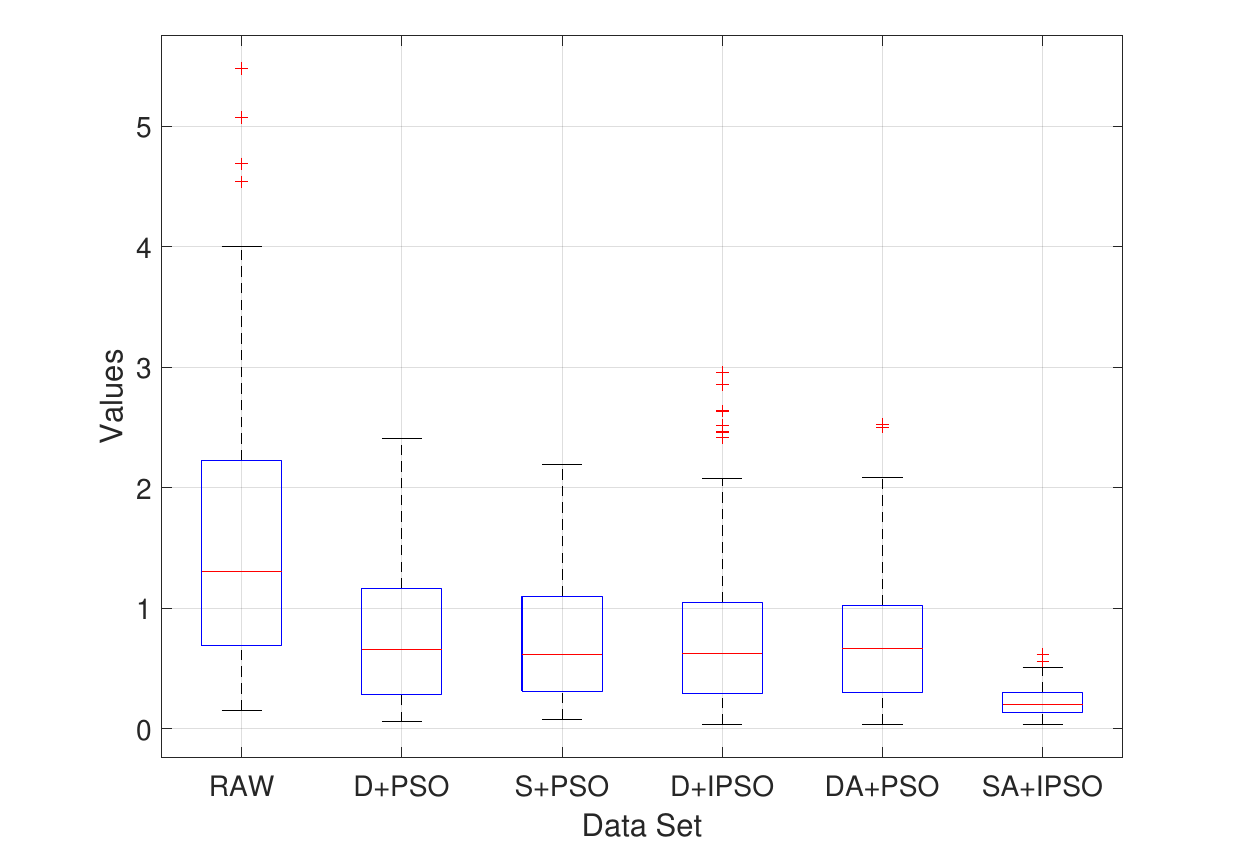}
		\caption{Boxplots of localization errors for the five compared methods (dual-UAV scenario).}
		\label{fig:dual_box}
	\end{figure}
	
	\begin{table}[htb]
		\centering
        \small
		\caption{Statistical comparison of localization errors for dual-UAV scenario (meters)}
		\label{tab:err_dual}
		\begin{tabular}{@{}lccccccc@{}}
			\toprule
			Method & Median & Mean & Std Dev & Outliers & Max & Min \\
			\midrule
			Raw & 1.31 & 1.60 & 1.16 & 4 & 5.48 & 0.15 \\
			D+PSO & 0.66 & 0.77 & 0.57 & 0 & 2.41 & 0.06 \\
			S+PSO & 0.61 & 0.73 & 0.52 & 0 & 2.20 & 0.07 \\
			D+IPSO & 0.63 & 0.79 & 0.67 & 7 & 2.96 & 0.03 \\
            DA+PSO & 0.66 & 0.75 & 0.56 & 2 & 2.53 & 0.03 \\
            SA+IPSO & 0.20 & 0.22 & 0.12 & 2 & 0.62 & 0.04 \\
			\bottomrule
		\end{tabular}
	\end{table}
    
    The ablation study results in Table~\ref{tab:err_dual} reveal important synergistic interactions among the proposed components in the dual-UAV setting. Comparing \textbf{S+PSO} against \textbf{D+PSO}, the spectrally-weighted objective function with standard PSO achieves a modest improvement: median error is reduced by \textbf{7.58\%} and mean error by \textbf{5.19\%}. This demonstrates that the spectral weighting provides some benefit in the dual-UAV context, likely due to its stabilizing effect on the multi-objective function landscape, but the improvement is limited without the enhanced optimizer.

    Comparing \textbf{D+IPSO} against \textbf{D+PSO}, the improved PSO alone with conventional D-optimality yields mixed results: the median error is reduced by \textbf{4.55\%}, but the mean error actually increases by \textbf{2.60\%}, and the number of outliers rises from 0 to 7. This indicates that in multi-UAV scenarios, the motion-constrained optimizer without proper geometric coordination can destabilize the planning process. The particle normalization, when applied to the FIM objective function alone, may overemphasize the information gain ignoring the triangulation geometry, leading to poor geometric configurations occasionally.

    Comparing \textbf{DA+PSO} against \textbf{D+PSO}, adding the intersection angle sine objective function to the conventional D-opt + standard PSO yields no improvement in median error (0.00\% reduction) and only a marginal \textbf{2.60\%} reduction in mean error. This is a critical finding: the intersection angle objective function, when paired with the conventional PSO and D-optimality, cannot effectively guide the swarm toward favorable triangulation geometries because the standard optimizer lacks the motion constraints to execute the required maneuvers and the normalization to balance competing objective functions.

    Finally, comparing \textbf{SA+IPSO} (the full proposed method) against \textbf{D+PSO}, the integration of all three contributions---spectrally-weighted FIM, intersection angle sine, and improved PSO---achieves a \textbf{69.70\%} reduction in median error and a \textbf{71.43\%} reduction in mean error. Relative to \textbf{S+PSO} (spectral weighting alone), \textbf{SA+IPSO} achieves an additional \textbf{67.21\%} median reduction, demonstrating that the intersection angle objective function and the improved optimizer are essential for realizing the full potential of the spectral weighting. Relative to \textbf{DA+PSO} (intersection angle with conventional optimizer), \textbf{SA+IPSO} achieves a \textbf{69.70\%} median reduction, confirming that the intersection angle sine objective function requires the improved PSO to effectively steer the dual-UAV formation toward orthogonal viewing geometries.
	
	\subsubsection{Axial Error Distribution Analysis}
	\label{sec:iso_dual}

\begin{figure*}[htb]
		\centering
		\includegraphics[width=0.9\textwidth]{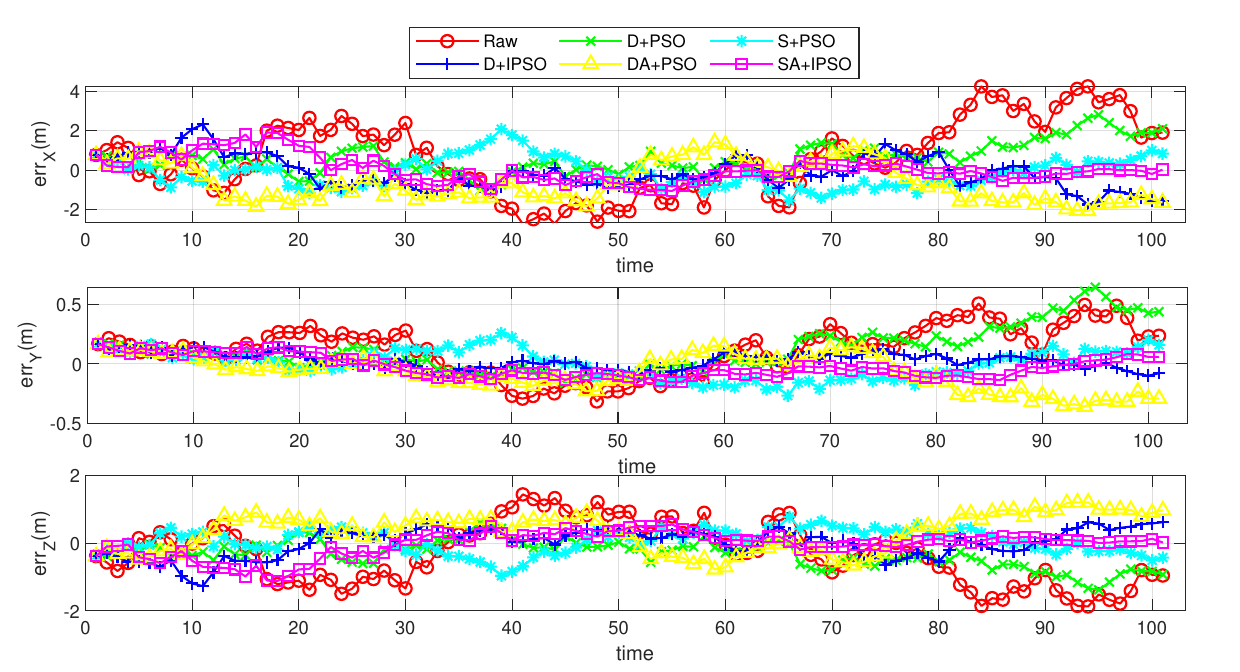}
		\caption{Localization error decomposition along x , y , and z  axes for dual-UAV scenario.}
		\label{fig:dual_iso}
	\end{figure*}

    Following the approach in Section~\ref{sec:iso_single}, the target localization errors along the $x$, $y$, and $z$ axes were extracted from the 100 simulation trials. Fig.~\ref{fig:dual_iso} illustrates the localization errors along three orthogonal axes for a representative trial. The results indicate that target localization errors following trajectory planning via our method are more uniformly distributed.
    
\begin{figure}[htb]
		\centering
		\includegraphics[width=0.49\textwidth]{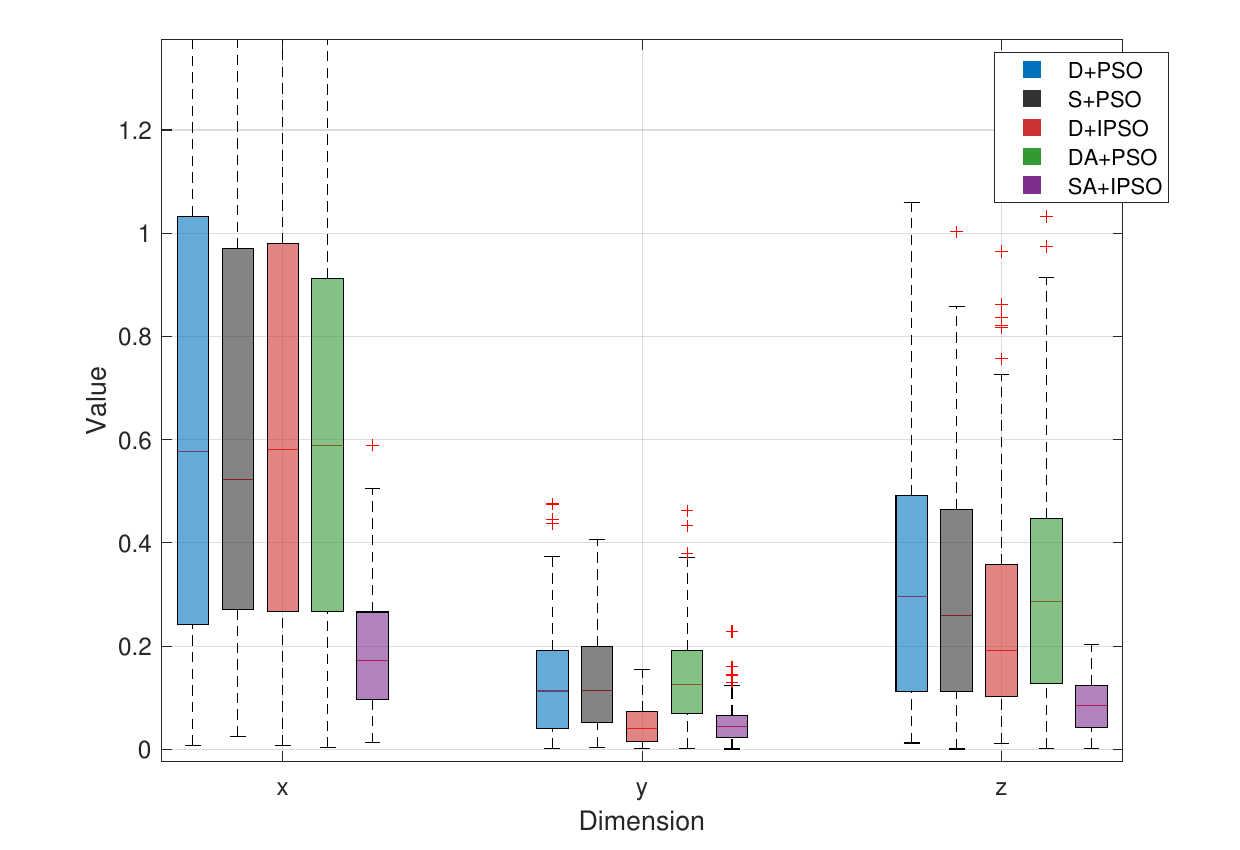}
		\caption{Box plots of localization error components for dual-UAV observers scenario.}
		\label{fig:dual_iso_box}
	\end{figure}

    To mitigate stochastic effects, error components along the $x$, $y$, and $z$ axes from 100 trials were statistically analyzed as boxplots. Fig.~\ref{fig:dual_iso_box} and Table~\ref{tab:iso_dual} demonstrate the differences in error distribution across three directions.
	
	\begin{table}[tb]
		\centering
		\caption{Axial error distribution comparison for dual-UAV scenario (meters)}
		\label{tab:iso_dual}
		\resizebox{\columnwidth}{!}{
			\begin{tabular}{@{}llcccccc@{}}
				\toprule
				Axis & Method & Median & Mean & Std Dev & Outliers & Max & Min \\
				\midrule
				\multirow{5}{*}{$x$} & D + PSO & 0.58 & 0.68 & 0.51 & 0 & 2.11 & 0.01 \\
				& S + PSO & 0.52 & 0.64 & 0.46 & 0 & 1.91 & 0.02 \\
				& D + IPSO & 0.58 & 0.74 & 0.64 & 7 & 2.79 & 0.01 \\
                & DA + PSO & 0.59 & 0.66 & 0.49 & 2 & 2.25 & 0.003 \\
                & SA + IPSO & 0.17 & 0.19 & 0.12 & 1 & 0.59 & 0.01 \\
				\midrule
                \multirow{5}{*}{$y$} & D + PSO & 0.11 & 0.13 & 0.11 & 3 & 0.48 & 0.001 \\
				& S + PSO & 0.11 & 0.13 & 0.09 & 0 & 0.41 & 0.004 \\
				& D + IPSO & 0.04 & 0.05 & 0.04 & 0 & 0.15 & 0.001 \\
                & DA + PSO & 0.12 & 0.14 & 0.10 & 3 & 0.46 & 0.002 \\
                & SA + IPSO & 0.05 & 0.05 & 0.04 & 5 & 0.23 & 0.001 \\
				\midrule
                \multirow{5}{*}{$z$} & D + PSO & 0.30 & 0.33 & 0.25 & 0 & 1.06 & 0.01 \\
				& S + PSO & 0.26 & 0.31 & 0.23 & 1 & 1.00 & 0.001 \\
				& D + IPSO & 0.19 & 0.26 & 0.22 & 6 & 0.96 & 0.01 \\
                & DA + PSO & 0.29 & 0.32 & 0.26 & 3 & 1.11 & 0.001 \\
                & SA + IPSO & 0.08 & 0.09 & 0.05 & 0 & 0.20 & 0.002 \\
				\bottomrule
			\end{tabular}
		}
	\end{table}
	
    The results exhibit distinct characteristics across the methods. The conventional FIM method (\textbf{D+PSO}) achieves higher accuracy along the $y$-axis but suffers from significant accuracy degradation in the other two dimensions. \textbf{S+PSO} (spectral weighting alone) shows modest balanced improvements. \textbf{D+IPSO} (improved PSO alone) achieves the best $y$-axis accuracy (median 0.04~m) but produces numerous outliers in $x$ and $z$ (7 and 6, respectively), indicating unstable geometric coordination. \textbf{DA+PSO} (intersection angle with conventional optimizer) actually degrades $y$-axis accuracy relative to D + PSO. In contrast, \textbf{SA+IPSO} (the full proposed method) appropriately relaxes the $y$-axis accuracy requirements to achieve balanced improvements across all three dimensions, with median errors of 0.17~m ($x$), 0.05~m ($y$), and 0.08~m ($z$), most lower than the baseline.

    For dual-UAV observation, the proposed full method achieves a median accuracy of 0.20~m for 1100~m range targets, representing a \textbf{69.70\%} improvement over conventional FIM methods. Furthermore, the proposed method exhibits more balanced axial error characteristics: while conventional FIM methods achieve high accuracy along the UAV motion direction but poor accuracy in other directions, the proposed method effectively improves localization accuracy across all directions simultaneously.
	
	\section{Discussion}
	\label{sec:discussion}
	
	This paper proposes novel trajectory planning methods based on conventional FIM optimization. To validate feasibility and theoretical performance improvements, simulations were designed for single-UAV and dual-UAV observation scenarios of a single moving target. Comparisons were conducted among \textbf{Raw} (no planning), D-optimality FIM with intersection angle (\textbf{D+PSO}), spectrally-weighted FIM with standard PSO (\textbf{S+PSO}), D-optimality FIM with improved PSO (\textbf{D+IPSO}), D-optimality FIM and intersection angle with standard PSO (\textbf{DA+PSO}), spectrally-weighted FIM with improved PSO (\textbf{S+IPSO}), and spectrally-weighted FIM and intersection angle with improved PSO (\textbf{SA+IPSO}).
	
	Experimental results demonstrate that for single-UAV observation, the proposed method achieves approximately 5~m terminal prediction point median distance error after planning, representing a \textbf{99.21\%} reduction compared to the conventional FIM method. For dual-UAV observation, terminal prediction point median distance error is reduced by \textbf{69.70\%}. The trajectory optimization algorithm exhibits superior performance in bearing-only localization tasks involving long-range, high-maneuverability targets.
	
    The ablation study design provides valuable insight into the individual and synergistic contributions of each proposed component. In the single-UAV scenario, the improved PSO emerges as the dominant factor: \textbf{D+IPSO} (improved PSO with D-optimality) achieves 99.16\% error reduction, while \textbf{S+PSO} (spectral weighting with standard PSO) shows virtually no improvement. This suggests that in single-observer settings, the primary bottleneck is optimizer feasibility and stability rather than the FIM criterion itself. The spectrally-weighted objective function contributes marginal additional gain (6.28\% over \textbf{D+IPSO}) by refining local search near degenerate boundaries.

    In the dual-UAV scenario, the story is more nuanced. Here, no single component alone achieves substantial improvement: \textbf{S+PSO} provides modest gain (7.58\%), \textbf{D+IPSO} yields mixed results with increased outliers, and \textbf{DA+PSO} shows no median improvement at all. Only when all three components are integrated (\textbf{SA+IPSO}) does the method realize its full potential. This strong synergy arises because the intersection angle objective function demands precise geometric coordination that the standard PSO cannot deliver, while the improved PSO requires a geometric guidance term to prevent the normalization scheme from over-emphasizing FIM at the expense of triangulation quality. The spectrally-weighted FIM stabilizes the multi-objective function landscape, ensuring that neither term dominates.
	
	However, one potential limitation of our method is its requirement for adequate prior observation information to guarantee convergence. As described in Section~\ref{sec:feas_single}, the UAV must first track the target along a predetermined trajectory for a period to acquire sufficient information. Without adequate prior observations, convergence requires significantly longer duration. 
	
	Second, in single-UAV observation scenarios, the target must not undergo drastic motion changes within short time intervals, otherwise localization accuracy may rapidly diverge. This fundamental limitation arises because the proposed method employs Trajectory Intersection to obtain target motion parameters within short time windows. If the target undergoes abrupt motion changes, the Trajectory Order may change abruptly, leading to convergence failure.
	
	Future work will investigate adaptive weight adjustment mechanisms for the multi-objective function and extend the framework to heterogeneous UAV swarms with mixed sensor modalities. Additionally, the development of online initialization strategies that reduce the required prior observation window would further enhance the practical applicability of the method.
	
	\section{Conclusion}
	\label{sec:conclusion}
	
	This paper presents an online trajectory optimization method for bearing-only localization using single and dual UAV platforms in dynamic environments. Beyond conventional FIM considerations, the method incorporates an intersection angle sine objective function for dual-UAV coordination and a spectrally-weighted FIM objective function that provides stronger gradients near degenerate configurations. The planning method improves upon conventional Particle Swarm Optimization by incorporating platform motion model constraints to prevent trajectory discontinuities and ensure physical feasibility. For objective function weighting, a particle normalization adaptive scheme is proposed based on PSO characteristics, addressing trajectory divergence under data singularity conditions.
	
	Simulation results demonstrate that for single-UAV scenarios, target trajectories are reasonably predicted, with terminal prediction point median distance error reduced by \textbf{99.21\%} after planning. For dual-UAV scenarios, terminal prediction point median distance error is reduced by \textbf{69.70\%}. Our method exhibits superior performance in long-duration bearing-only target localization of maneuverability targets at extended ranges.
	
	\bibliography{references}
    \end{sloppypar}
\end{document}